\documentclass[10pt, conference, compsocconf]{IEEEtran}
\usepackage{amsfonts}

\ifCLASSINFOpdf
   \usepackage[pdftex]{graphicx}
   \DeclareGraphicsExtensions{.pdf,.jpeg,.png}
\else
\fi
\usepackage[cmex10]{amsmath}
\newcommand{\norm}[1]{\left\lVert#1\right\rVert}

\DeclareMathOperator*{\argmin}{arg\,min}

\usepackage[caption=false,font=footnotesize]{subfig}
\usepackage{booktabs}

\usepackage{color}
\definecolor{turquoise}{cmyk}{0.65,0,0.1,0.3}
\definecolor{purple}{rgb}{0.65,0,0.65}
\definecolor{dark_green}{rgb}{0, 0.5, 0}
\definecolor{orange}{rgb}{0.8, 0.6, 0.2}
\definecolor{red}{rgb}{0.8, 0.2, 0.2}
\definecolor{darkred}{rgb}{0.6, 0.1, 0.05}
\definecolor{blueish}{rgb}{0.0, 0.3, .6}
\definecolor{light_gray}{rgb}{0.7, 0.7, .7}
\definecolor{pink}{rgb}{1, 0, 1}
\definecolor{greyblue}{rgb}{0.25, 0.25, 1}
\definecolor{gold}{rgb}{0.7, 0.5, 0}
\definecolor{gray}{gray}{0.5}

\newcommand{\comment}[1]{}

\usepackage{booktabs}

\usepackage{hyperref}

\newcommand{\Figure}[1]{Figure~\ref{fig:#1}}

\newcommand{\Table}[1]{Table~\ref{tab:#1}}

\newcommand{\Eq}[1]{Eq.~\ref{eq:#1}}

\newcommand{\Section}[1]{Section~\ref{sec:#1}}

\newcommand{\loss}[1]{{\mathcal{L}_{\text{#1}}}}

\newcommand{\IR}{{\mathbb{R}}}
\newcommand{\regressor}{{\mathcal{J}}}
\newcommand{\smpl}{{\boldsymbol{\Phi}}}
\newcommand{\pose}{{\boldsymbol{\Theta}}}
\newcommand{\shape}{{\boldsymbol{\beta}}}
\newcommand{\poseNshape}{{\boldsymbol{\theta}}}
\newcommand{\vertices}{{\boldsymbol{V}}}
\newcommand{\gtpoints}{{\mathbf{X}}}
\newcommand{\maskrcnn}{{\mathbf{S}}}
\newcommand{\render}{{\mathcal{R}}}
\newcommand{\discrim}{{\mathcal{D}}}
\newcommand{\paramdiscrim}{{\boldsymbol{\Psi}}}

\def \customparskip {0.3em}
\renewcommand{\paragraph}[1]{\vspace{\customparskip}\noindent\textbf{#1.}}

\begin{document}
%
\title{A Simple Method to Boost Human Pose Estimation Accuracy by Correcting\\ the Joint Regressor for the Human3.6m Dataset}


\author{\IEEEauthorblockN{Eric Hedlin}
\IEEEauthorblockA{
University of British Columbia\\
Vancouver, Canada\\
iamerich@cs.ubc.ca}
\and
\IEEEauthorblockN{Helge Rhodin}
\IEEEauthorblockA{
University of British Columbia\\
Vancouver, BC\\
rhodin@cs.ubc.ca}
\and
\IEEEauthorblockN{Kwang Moo Yi}
\IEEEauthorblockA{
University of British Columbia\\
Vancouver, BC\\
kmyi@cs.ubc.ca}
}


%


\maketitle

\begin{abstract}

Many human pose estimation methods estimate Skinned Multi-Person Linear (SMPL) models and regress the human joints from these SMPL estimates.
In this work, we show that the most widely used SMPL-to-joint linear layer (joint regressor) is inaccurate, which may mislead pose evaluation results.
To achieve a more accurate joint regressor, we propose a method to create pseudo-ground-truth SMPL poses, which can then be used to train an improved regressor. 
Specifically, we optimize SMPL estimates coming from a state-of-the-art method so that its projection matches the silhouettes of humans in the scene, as well as the ground-truth 2D joint locations.
While the quality of this pseudo-ground-truth is challenging to assess due to the lack of actual ground-truth SMPL, with the Human 3.6m dataset, we qualitatively show that our joint locations are more accurate and that our regressor leads to improved pose estimations results on the test set without any need for retraining.
We release our code and joint regressor at \url{https://github.com/ubc-vision/joint-regressor-refinement}

\end{abstract}

\begin{IEEEkeywords}
joint regressor; pseudo-ground-truth; silhouette; human body model
\end{IEEEkeywords}
\section{Introduction}
\label{sec:intro}

Human pose estimation is a fundamental and challenging computer vision problem. 
It is the task of estimating the human body pose from an image or a series of images. 
Traditional methods for human pose estimation~\cite{dantone2013human, gkioxari2014using} rely on
handcrafted feature detectors that are carefully designed for robustness.
Works include \cite{sminchisescu2002human}, which estimates poses by taking into consideration the subject's silhouette. 
More recently, deep learning-based methods
have dominated the field~\cite{hmr, smpl, vibe, spin} because of their superior performance. 
Well-designed network architectures, richer datasets, and more accurate and practical body models have since led to improved pose estimation accuracy~\cite{smpl, human3.6m, survey}.

In this paper, we work to improve the performance of the current state of the art.
We focus on the joint regressor, which estimates joint locations of humans ---the standard metric for measuring pose accuracy---from the Skinned Multi-Person Linear (SMPL) human body model~\cite{smpl}. 
The joint regressor is necessary for the generality of SMPL's use across different datasets since different labelled human pose datasets use ground truth joints associated with different body parts and the exact position of the same joint varies. 
A different joint regressor can be found per dataset, allowing supervision from multiple datasets for one model, whether it is a dataset with ground truth SMPL as in 3DPW \cite{3dpw}, ground truth 3D and 2D joints, as in Human3.6m or Total Capture \cite{total_capture, human3.6m}, or labelled (or even unlabeled) 2D data \cite{2ddataset, spin}.
An interesting empirical observation that we present in \Section{jreg_results} reveals that the joints regressed on the Human3.6m~\cite{human3.6m} dataset using the currently accepted joint regressor~\cite{spin, vibe} do not lead to accurate joint estimates.
This is a critical bottleneck of existing methods that utilize this model~ \cite{hmr, spin, vibe}, 
as this means that these methods will have added error in their joint estimates, eventually being misled about which methods are actually more accurate.

\begin{figure}%
    \centering
    \subfloat[\centering Initial Image]{
    {
        \label{subfig:initial_image}
\includegraphics[width=0.2\textwidth]{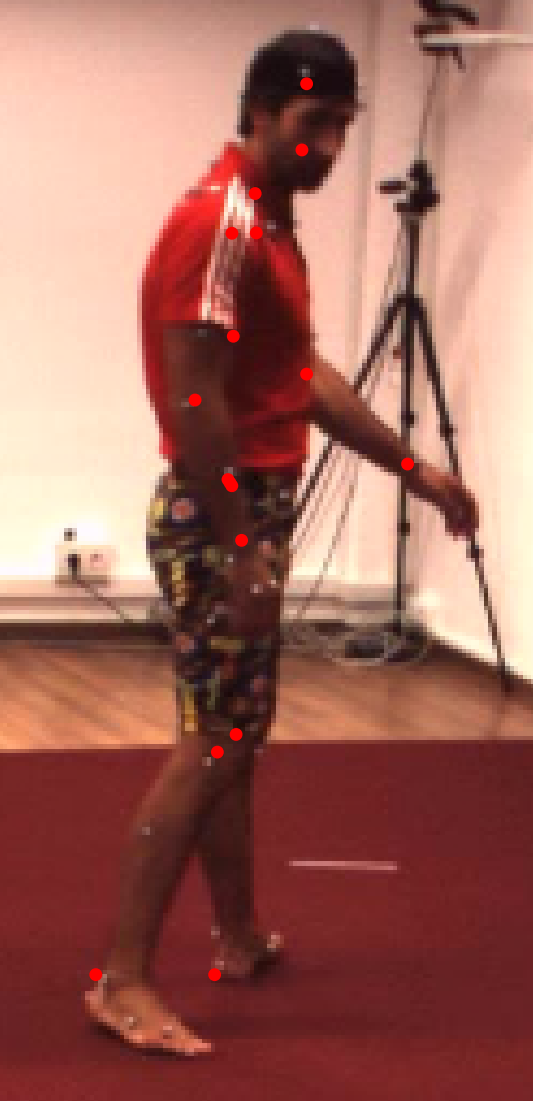} 
    }
    }%
    \subfloat[\centering Silhouette as estimated by Mask-RCNN]{
    {
        \label{subfig:maskrcnn_mask}
        \includegraphics[width=0.2\textwidth]{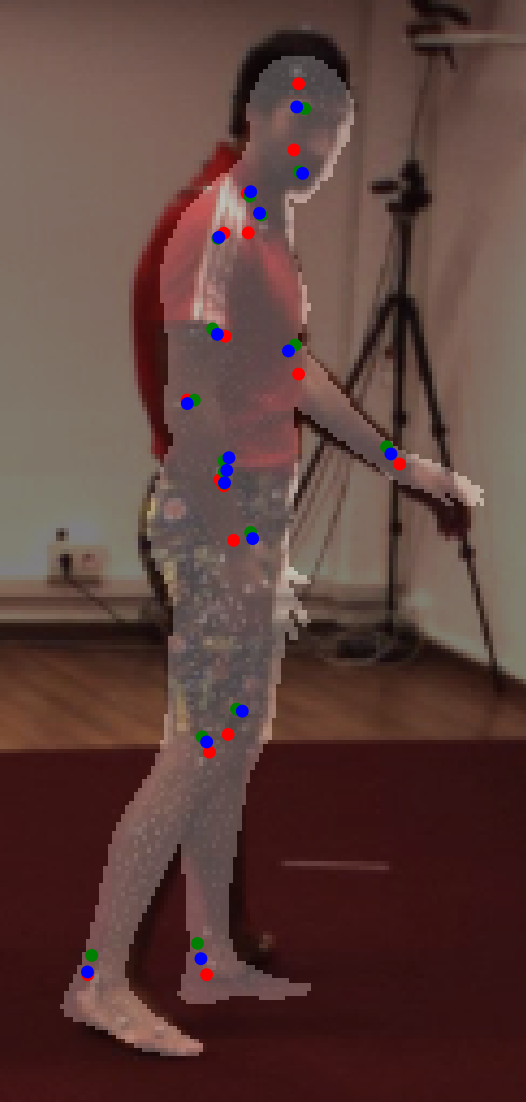}
    } 
    }%
    \caption{%
    {\bf Teaser -- }
    Example joint estimation from the same estimated SMPL pose with with \textit{different} SMPL-to-joint regressors.
    (left) ground truth joints projected onto the image, and (right) comparisons between ground truth joint locations, the commonly used joint regressor, and our regressor.
    Ground-truth joint locations are marked with \textcolor{red}{red}, projected joints from the commonly used joint regressor $\mathcal{J}$ in \textcolor{dark_green}{green}, and projected joints from our joint regressor $\hat{\mathcal{J}}$ in  \textcolor{blue}{blue}.
    Simply changing to our regressor leads to improved pose estimates.
    }%
    \label{fig:teaser}%
\end{figure}

A potential root-cause of this erroneous joint regressor stems from how this regressor was obtained.
The pseudo-ground-truth SMPL poses for fitting the currently accepted joint regressor on the Human3.6m dataset~\cite{human3.6m, IonescuSminchisescu11} were found using Motion and Shape Capture from Sparse Markers (MoSh)~\cite{mosh}.
MoSh takes the Shape Completion and Animation of People (SCAPE) human body model~\cite{scape} and a series of ground truth joints \cite{human3.6m} and estimates a set of plausible poses per frame through knowledge of constraints of how the human body moves. 
Notice how this process does not involve images from the dataset---it does not ensure that the estimated SMPL poses well align with what is being observed.
Thus, while the poses may be plausible, there is room for misalignment with the image. 
In fact, as we will show later in \Figure{opt_without_silhouette}, without considering both the plausiblity and the alignment together, at the very least in our case, we were unable to obtain SMPL meshes that satisfy both.

Hence, we first estimate pseudo-ground-truth SMPL human body parameters, constraining the poses to be biologically plausible and conforming to the person's silhouette.
We then use these estimates to retrain a new joint regressor.
Specifically, to create our pseudo ground truth, we start from SMPL estimates from SPIN~\cite{spin} and enhance these estimates via optimization.
We aim to integrate two constraints: plausibility and silhouette alignment.
For plausibility, we assume that the original SMPL poses from SPIN are plausible poses since SPIN has been trained with various human poses, and aim to not deviate too much from the original distribution of poses.
In other words we train a discriminator that discriminates between original SMPL poses and the optimized ones, and minimizes the discrepancy between the two distributions while optimizing.
For the silhouette, we follow in the footsteps of classic literature and minimize the discrepancy between the estimated segmentation and the render.
We accomplish this using an off-the-shelf human segmentor~\cite{maskrcnn}, and perform differentiable rendering~\cite{pytorch3d} of the SMPL vertices and pass the energy term for minimizing the difference between the two back to the pose parameters.

We verify the efficacy of our method on the most widely used 3D pose estimation benchmark, the Human 3.6m dataset.
Qualitatively, our regressed joints align significantly better with the marker locations on the human subjects compared to the standard regressor.  
Quantitatively, we show that our joint regressor can immediately improve SMPL-based human pose estimators~\cite{spin, vibe} without any retraining. 
\section{Related works}
\label{sec:Human-Body-Models}

We provide a brief review of work on human pose estimation and discuss the standard joint regressor in more detail.

\subsection{Human pose estimation}

Human pose estimation can roughly be grouped into two; model-free and model-based.
Model-free methods directly predict the final set of vertices or joints that describe the human body, whereas model-based methods utilize a handcrafted human body model and predict its parameters. 
Model-free methods rely completely on data, whereas model-based methods utilize prior knowledge of how human bodies behave, abstracted manually by researchers.

\paragraph{Model-free methods}
In more detail, model-free methods do not employ any human body models when reconstructing a 3D human representation~\cite{metro, survey}.
The final joint locations, or any target aspect of the human body, can either be directly regressed from the image, as in MEsh TRansfOrmer (METRO), which directly estimates all vertices and joints directly using a transformer~\cite{metro}. 
Intermediate 2D estimates can also be "lifted" to 3D as a final estimation~\cite{survey, lifting}, which is useful in being able to leverage the availability of 2D data, however it tends to discard the initial image during the lifting process. 
For a more comprehensive review of model-free methods see \cite{survey}.

\paragraph{Model-based methods}
\label{sec:model_based_estimation}
Model-based methods, on the other hand, leverage the fact that we know what constraints should be applied to the human body already such as possible joint rotations or relative limb lengths.

These constraints can easily be represented by a human body model. \cite{total_capture, smpl, pavlakos2019expressive}. 
In our work on human pose refinement, we focus on model-based pose estimation, specifically the SMPL human body model~\cite{smpl}.
SMPL is a vertex-based linear model which can represent a broad range of shapes \cite{smpl}.

\paragraph{The SMPL model}
In this work, we focus on the SMPL model as it and its variants~\cite{smplx} are arguably the most popular.\footnote{Our work can be trivially applied to \textit{any} mesh-based human body model.}
The SMPL body model expresses the human body in terms of pose $\pose$ and shape parameters, which together parameterize the model $\poseNshape = (\pose, \shape)$.
The pose parameters $\pose \in \IR^{24\times3}$ denotes the rotations of the twenty-four joints relative to its parent, starting from a root joint (the pelvis joint). 
The shape parameters $\shape \in \IR^{10}$ are the ten Principle Component Axes (PCA) that define the shape of the human body, extracted from a set of possible human body shapes~\cite{smpl}. 
Denoting this model is $\smpl(\cdot)$, the output of the SMPL model $\smpl(\poseNshape)$ is a deformation of an artist-created mesh.
Thus, $\smpl(\poseNshape)$ is then a set of vertices $\vertices \in \IR^{6890 \times 3}$. 
Typically, methods that utilize SMPL and its variants use these vertices to regress the joint locations via linear regression.

\subsection{The standard joint regressor}
\label{sec:joint_regressor}

\begin{figure}
  \centering
    \includegraphics[width=.45\linewidth]{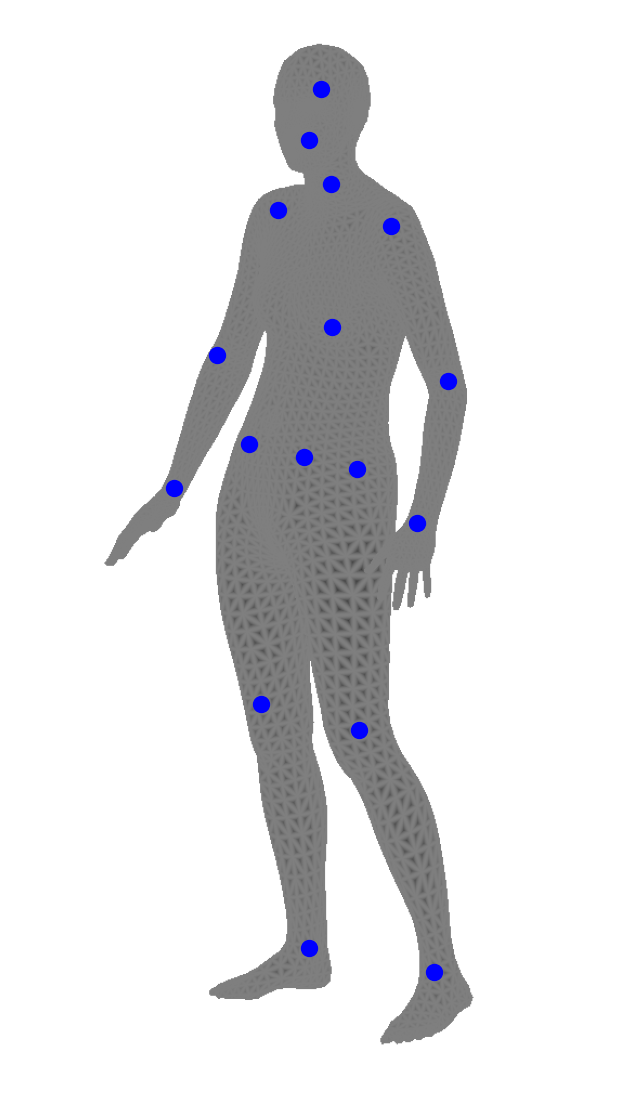}
    \includegraphics[width=.45\linewidth]{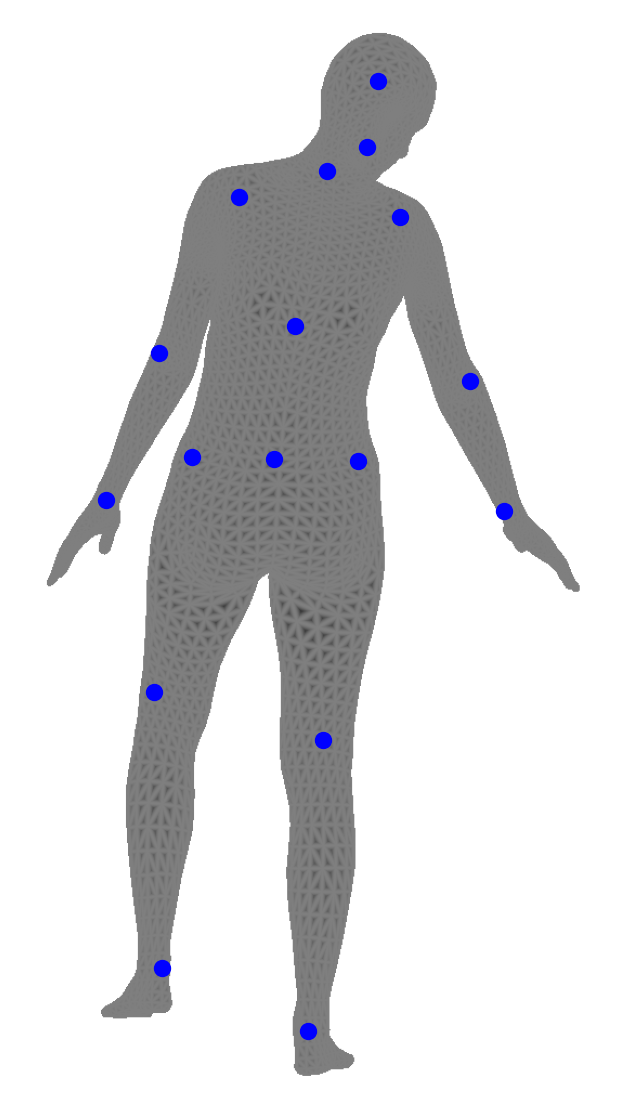}
  \caption{{\bf SMPL vertices and annotated joints -- }
    Example poses with visualizations of the regressed joints for the Human 3.6m dataset \cite{human3.6m}. The set of lines connecting the mesh of vertices are shown in \textcolor{gray}{gray} and the regressed joints are shown in \textcolor{blue}{blue}.
    }%
  \label{fig:joint_regressor}
\end{figure}

As can be seen in \Figure{joint_regressor}, a joint regressor distills the output set of vertices $\vertices$  down to a smaller set of points corresponding to the human joint locations, which can then be used, for example, to evaluate the quality of the mesh against human annotation. 
This joint regressor is a simple linear transformation, $\regressor \in \IR^{|J| \times 6890}$ where $J$ are the target joints. 
Regressed joint positions can then be found through a simple matrix multiplication between vertices and the regressor as $\Hat{J} = \regressor\vertices$.
In~\cite{smpl}, the joint regressor is found by optimizing $\regressor$ to match a set of vertices to known joint locations via non-negative least squares to encourage sparsity, a trait that is desirable when regressing the joints, since only few nearby vertices should determine the joint locations to allow for the regression to be robust against shape and pose changes of the human mesh.
The regression weights are also encouraged to sum to one so that the regressed joints are within the convex hull of the vertices.

In the case of the joint regressor that is typically used for the Human3.6m~\cite{human3.6m} dataset, as mentioned earlier in \Section{intro}, it's based on pseudo-ground-truth vertices from MoSh~\cite{mosh}.
MoSh estimates body shape, pose and soft tissue deformations directly from sparse markers, but it does not consider images. 
Thus, while the pseudo-ground-truth vertices may regress to the ground-truth 3D joints using the discovered regressor, nothing explicitly encourages the estimated vertices to align with the human subject. 
In this work, we show that by using pseudo ground truth that considers also the alignment, we can obtain an improved regressor that reveals that the SMPL-based methods \cite{vibe, spin, meva} are more accurate than they were previously thought to be.

We train the joint regressor in a style similar to that of SMPL, using a non-negative least
squares \cite{lawson1995solving} with the inclusion of a term that
encourages the weights to add to one,
\begin{equation}
    \mathcal{L}_{sum} = \sum_{j \in \regressor} \norm{\left(\sum_{i=1}^{|V|} J_{i, j} > 0\right) - 1}_2 \;.
    \label{eq:jointregressor}
\end{equation}
This encourages sparsity and discourages predicting joints outside the body. 
\section{Method}
\label{sec:jreg_method}
\label{sec:est_pseudo_gt_verts}

To obtain a better joint regressor, we start from SMPL estimates given by a state-of-the-art model, SPIN~\cite{spin}. We optimize the estimate's SMPL parameters $\poseNshape$, considering how plausible the pose is and how well the projected vertices align with the image and the ground-truth joint annotations.
Specifically, we find the optimized parameters $\poseNshape_\text{opt}$ according to the combination of three energy terms as
\begin{equation}
    \poseNshape_\text{opt} = \argmin_\poseNshape 
    \loss{joint}(\poseNshape) 
    + \loss{silhouette}(\poseNshape)
    + \loss{adv}(\poseNshape)
    ,
\end{equation}
which we detail in the following subsections.
We then use the optimized parameters $\poseNshape_\text{opt}$ to obtain a new joint regressor according to \Eq{jointregressor}.

\subsection{Ground-truth joint alignment -- $\loss{joint}$}

As we are aiming to generate pseudo ground truth, the first objective is for our estimated SMPL parameters $\poseNshape$ to result in proper 3D joint locations $\gtpoints$, once a joint regressor is used on top of the vertices $\smpl(\poseNshape)$.
To implement this objective,
we cannot rely on the standard joint regressor $\regressor_\text{std}$ as we have shown this will encourage improper fittings.
Instead we initialize a new joint regressor $\regressor_\text{spin}$ by fitting estimated poses from SPIN $\poseNshape$ to the ground truth 3D joint locations $\gtpoints$. 
Mathematically, given SMPL parameters $\poseNshape$, we write
\begin{equation}
    \loss{joint} =\norm{\gtpoints-\regressor_\text{spin}(\smpl(\poseNshape))}_2^2
    .
\end{equation}

\subsection{Aligning the mesh with the silhouette -- $\loss{silhouette}$}

\begin{figure}%
    \centering
    \subfloat[\centering Input image]{%
        \label{subfig:initial_image}
        \includegraphics[width=0.3\linewidth]{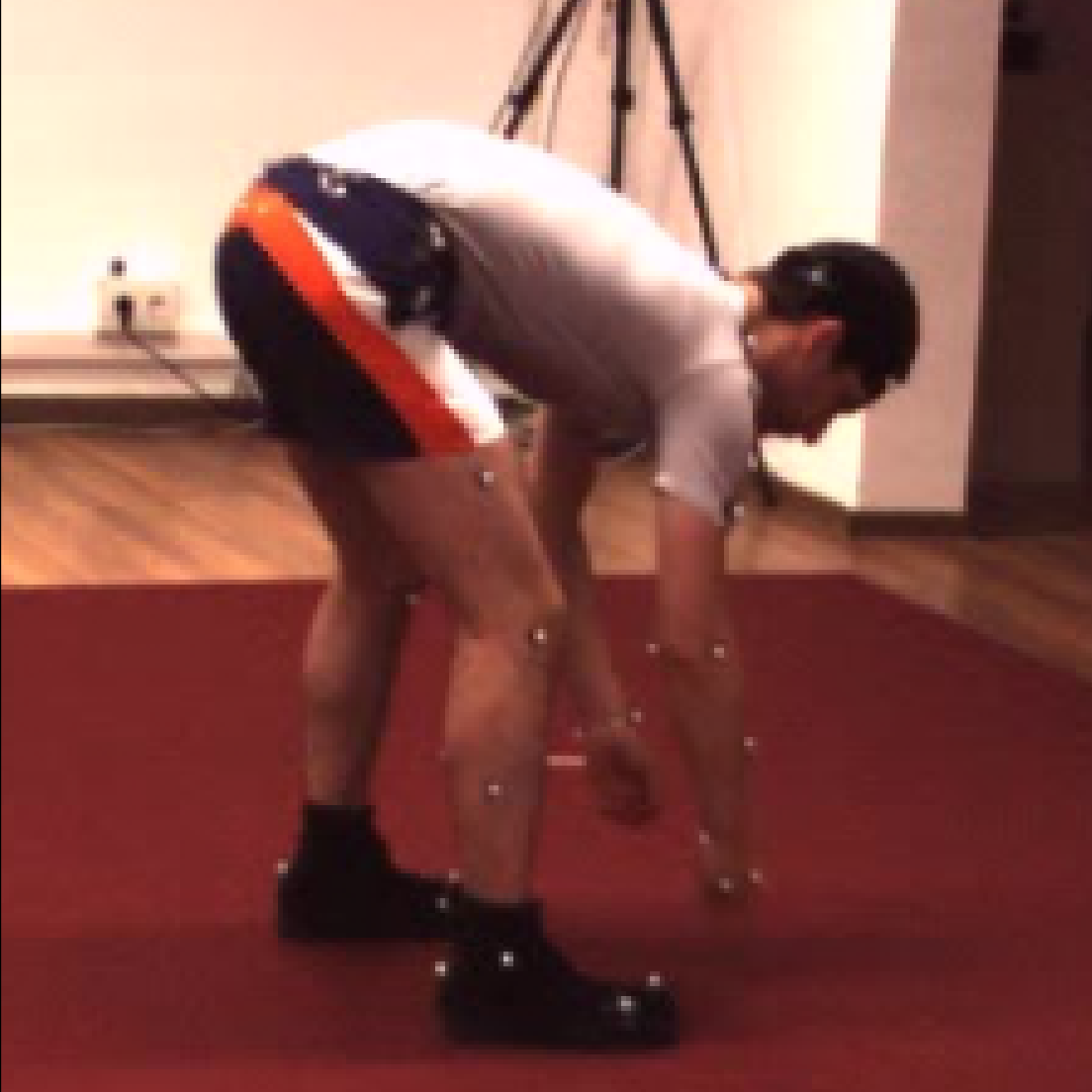} 
    }%
    \subfloat[\centering Mask R-CNN silhouette]{%
        \label{subfig:maskrcnn_mask}
        \includegraphics[width=0.3\linewidth]{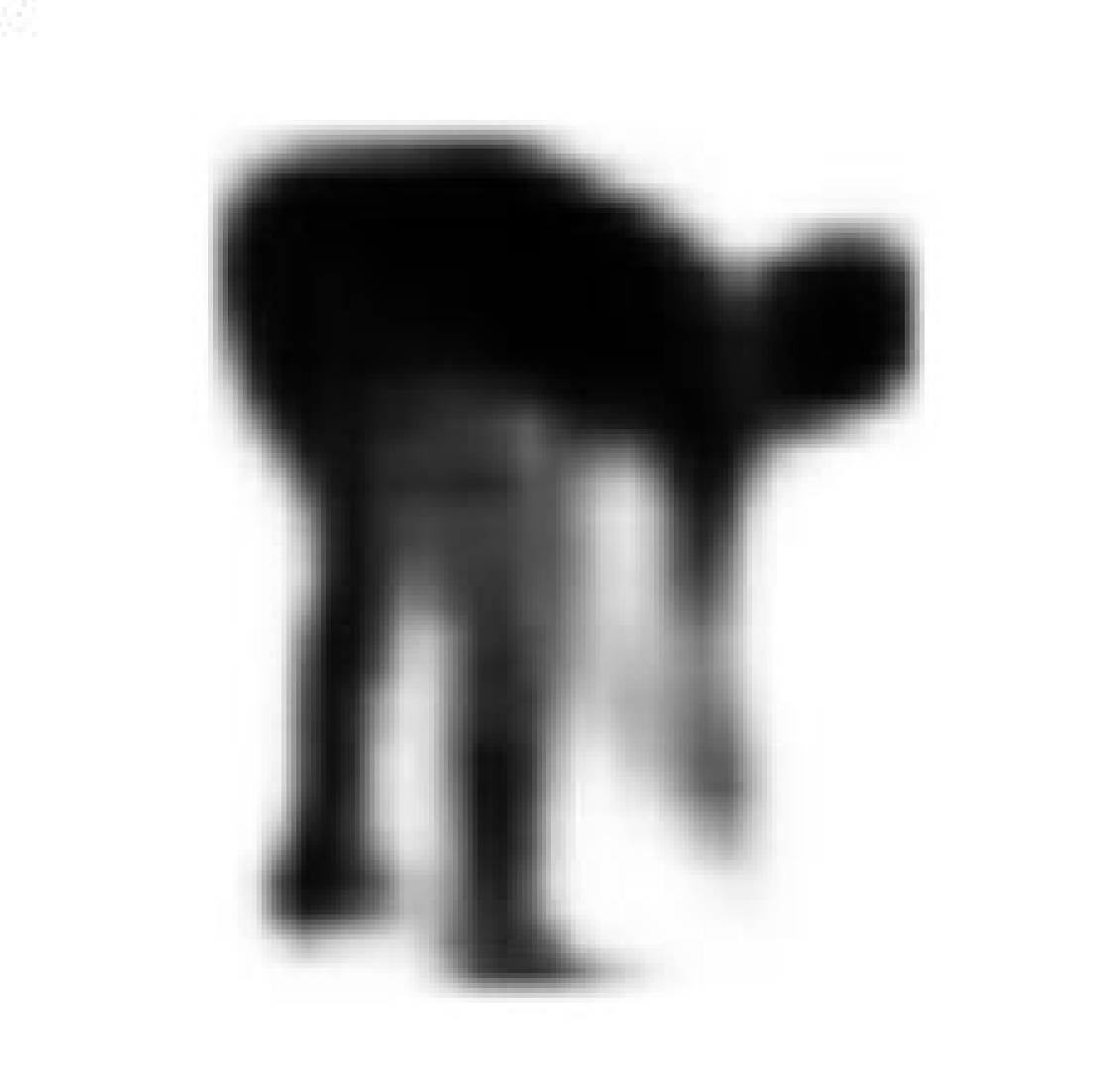}
    }%
    \subfloat[\centering Estimated silhouette]{%
        \label{subfig:est_silhouette}
        \includegraphics[width=0.3\linewidth]{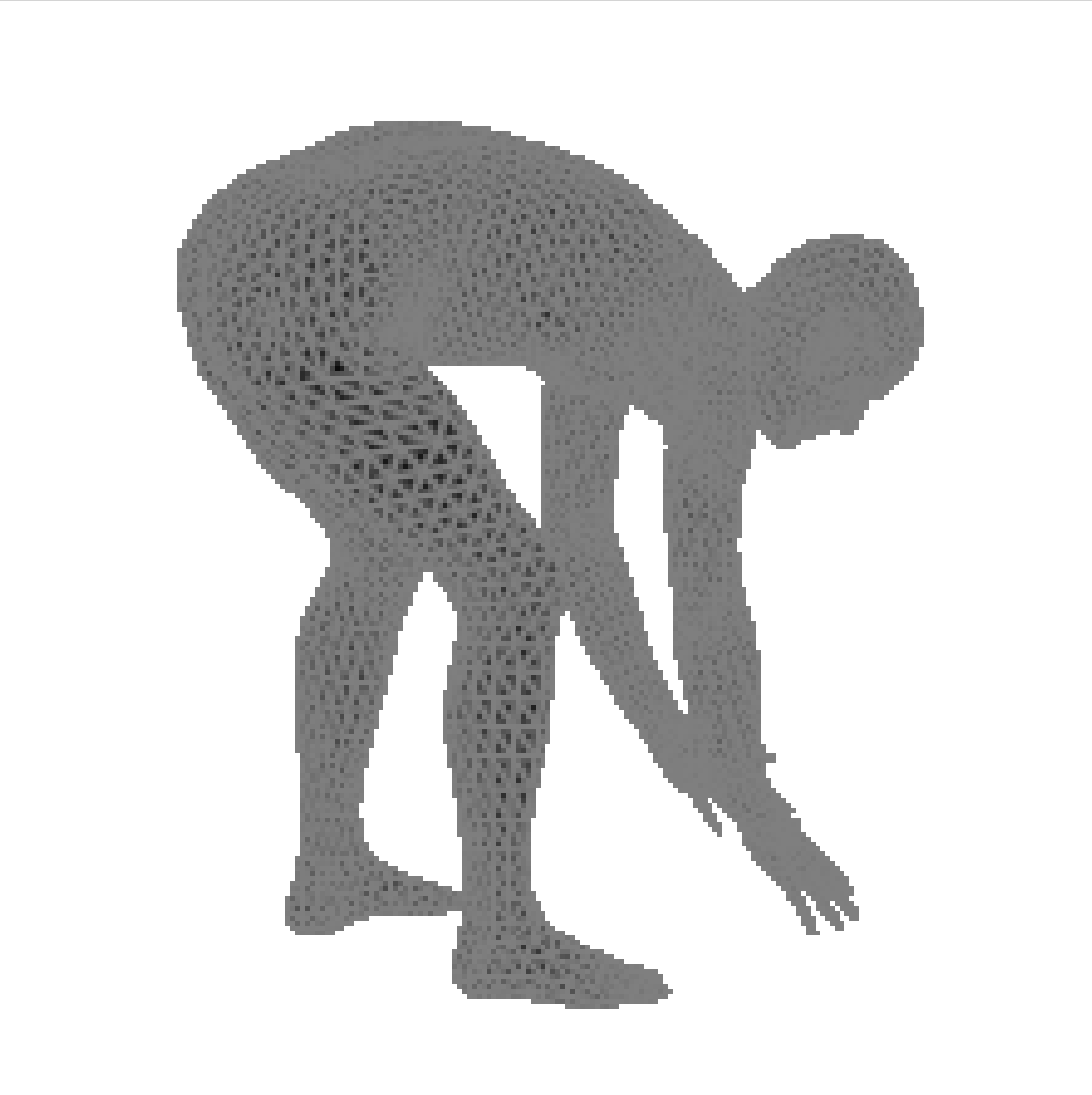} 
    }%
    \caption{{\bf Silhouette example -- }
    The input image is shown on the left. The corresponding Mask R-CNN estimate for a person is shown in the middle. The estimated pose from SPIN for the input image rendered using Pytorch3d \cite{pytorch3d} is shown on the right. 
    }%
    \label{fig:est_gt_silhouette}%
\end{figure}
To enforce our pseudo ground truth to align with the human subject in the image, we introduce an energy term that encourages the rendered pose to match the silhouettes of the human subject.
To obtain the pseudo ground truth silhouette of the human subject, we rely on an off-the-shelf human segmentor, specifically Mask R-CNN~\cite{maskrcnn}.
If we denote the mask provided by Mask R-CNN as $\maskrcnn$, we write
\begin{equation}
    \loss{silhouette} = \norm{\maskrcnn-\render(\smpl(\poseNshape))}_2^2
    ,
\end{equation}
where $\render$ is the differentiable rendering of the vertices into the image, which we implement via PyTorch 3D~\cite{pytorch3d}; see \Figure{est_gt_silhouette} for an example.

\subsection{Enforcing plausible poses -- $\loss{adv}$}

\begin{figure}%
    \centering
    \includegraphics[width=0.49\linewidth]{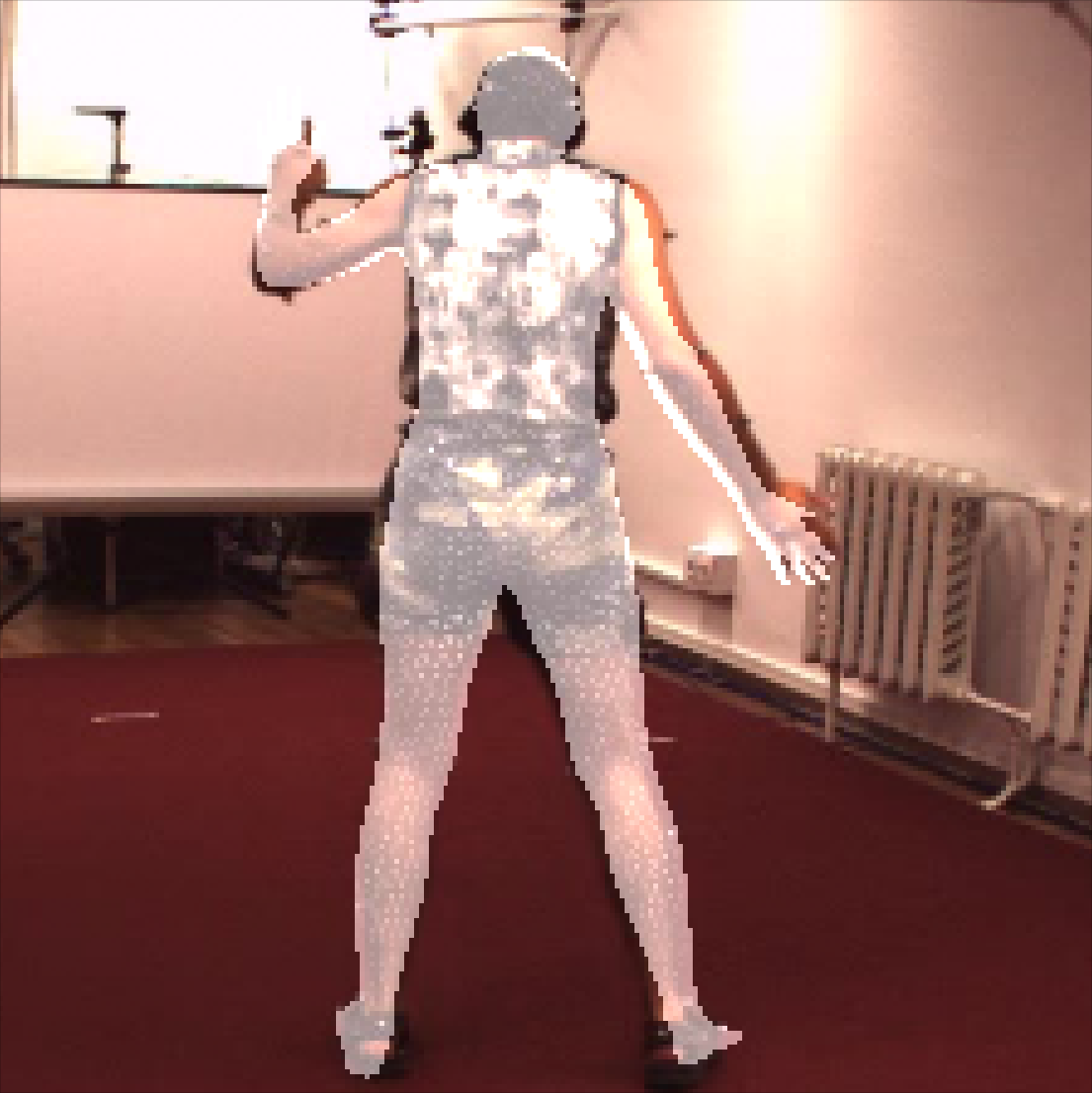}
    \includegraphics[width=0.49\linewidth]{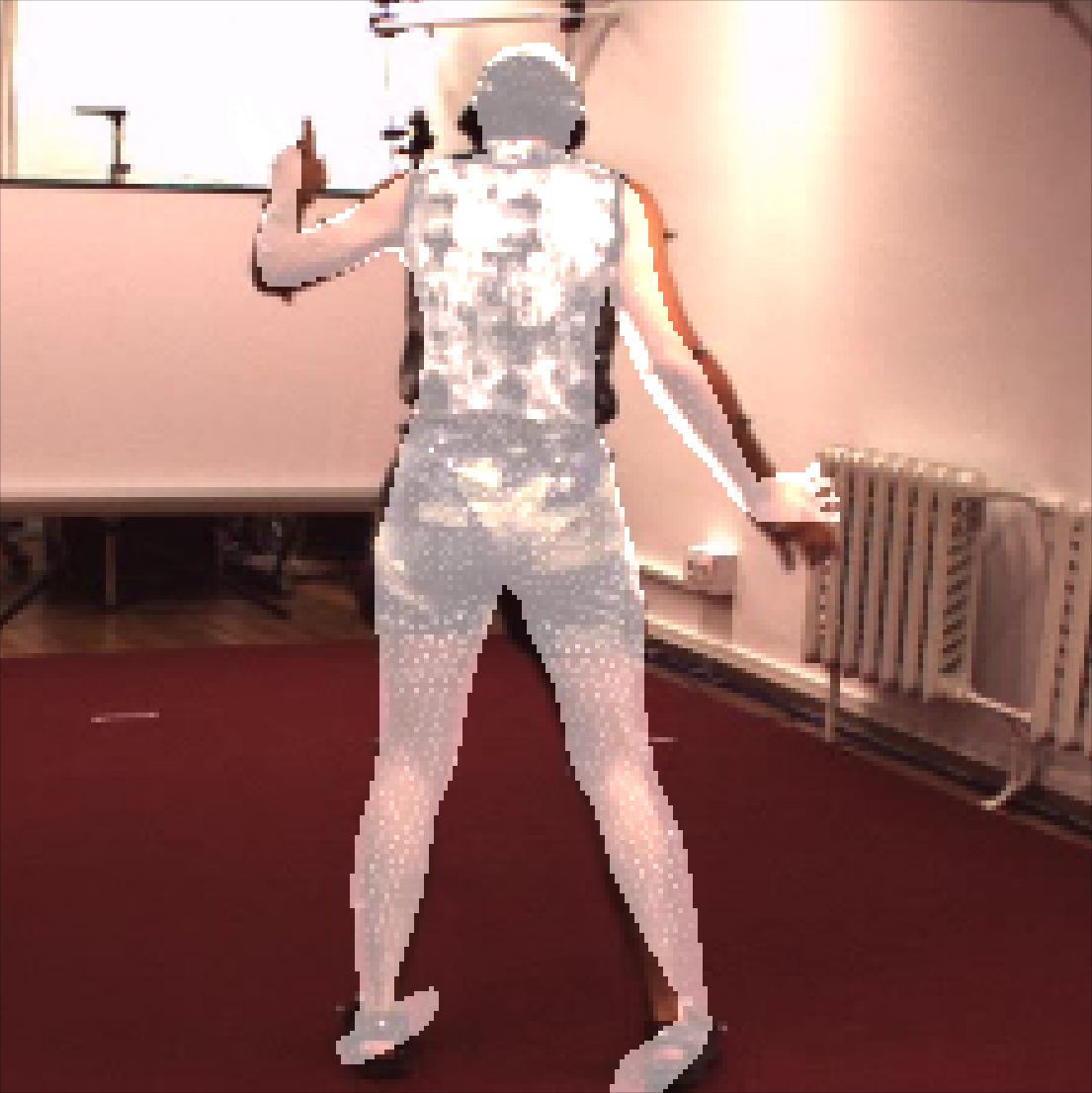}
    \caption{
    {\bf Effect of discriminator -- }
    We show the optimized SMPL parameters (left) with and (right) without the discriminator, overlaid onto the image.
    Having the discriminator encourages poses to be plausible, unlike the one without it, which focuses solely on getting the silhouette and the joints right, which may not be biologically plausible.
    Original Joint Regressor refers to the commonly accepted joint regressor.
    Retrained Joint Regressor refers to the joint regressor we trained simply matching SPIN outputs with ground truth joints. 
    Optimized Joint Regressor refers to the joint regressor refined on optimized poses. 
    }%
    \label{fig:ablation_discrim}%
\end{figure}

To enforce the poses of our pseudo ground truth to be plausible and realistic, we further introduce a
an objective based on an adversarial training setup~\cite{gan}.
Specifically, we train a discriminator network $\discrim_\paramdiscrim$ with parameters $\paramdiscrim$ that distinguishes between our optimized poses and shapes $\poseNshape_\text{opt}$ and the initial poses and shapes from SPIN $\poseNshape_\text{init}$, and optimize $\poseNshape_\text{opt}$ towards fooling $\discrim$.
In more detail, we find
\begin{equation}
    \paramdiscrim^*=\argmin_\paramdiscrim
    \left\|1-\discrim_\paramdiscrim\left(\poseNshape_\text{init}\right)\right\|_2^2
    +
    \left\|\discrim_\paramdiscrim\left(\poseNshape_\text{opt}\right)\right\|_2^2
    ,
\end{equation}
and define our energy term as
\begin{equation}
    \loss{adv} = \norm{1-\discrim_{\paramdiscrim^*}(\poseNshape_\text{opt})}_2^2
    .
\end{equation}
This term plays a critical role as shown in \Figure{ablation_discrim}. 

\subsection{Implementation details}

We followed the classic literature during joint regressor training and split the data into training and testing sets. Subjects S1, S5, S6, S7, and S8 are used for training and S9 and S11 are used for testing. 

The trained joint regressor is not used for optimizing poses as this could lead to joint drift. 
As a result, finding the optimized poses and training the joint regressor are done in two distinct steps. 

We optimize each pose for 100 iterations with a step length of 1e-2 and a momentum of 0.9 using Adam optimizer~\cite{adam}.
We weigh each energy term to be of approximately equivalent magnitude on the first iteration. 
The discriminator network $\discrim_\paramdiscrim$ weights are updated once each time a new pose is optimized. 
The discriminator is trained with Adam optimizer~\cite{adam}, a learning rate of 1e-3 and a momentum of 0.9.
This process is slow, but fortunately the network converges in a few hundred steps. 

While using the standard joint regressor would make our pseudo ground truth behave similar to the one MoSh provides, the other two terms $\loss{silhouette}$ and $\loss{adv}$ enforce our $\poseNshape_\text{opt}$ to deviate.

\section{Results}
\label{sec:jreg_results}

\def \qualfigwidth {0.16}
\begin{figure*}%
    \centering
    \includegraphics[width=\qualfigwidth\textwidth]{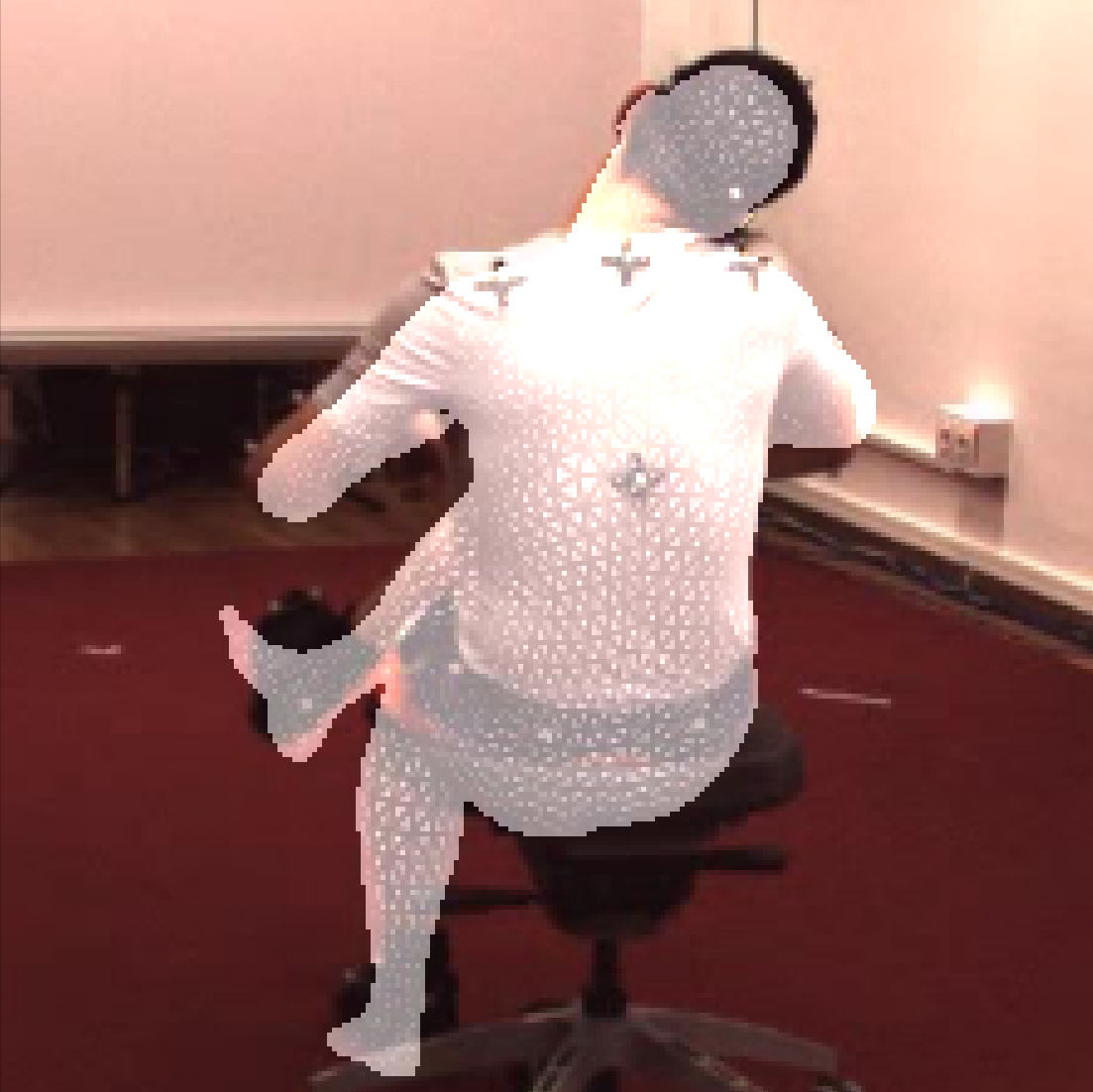}
    \includegraphics[width=\qualfigwidth\textwidth]{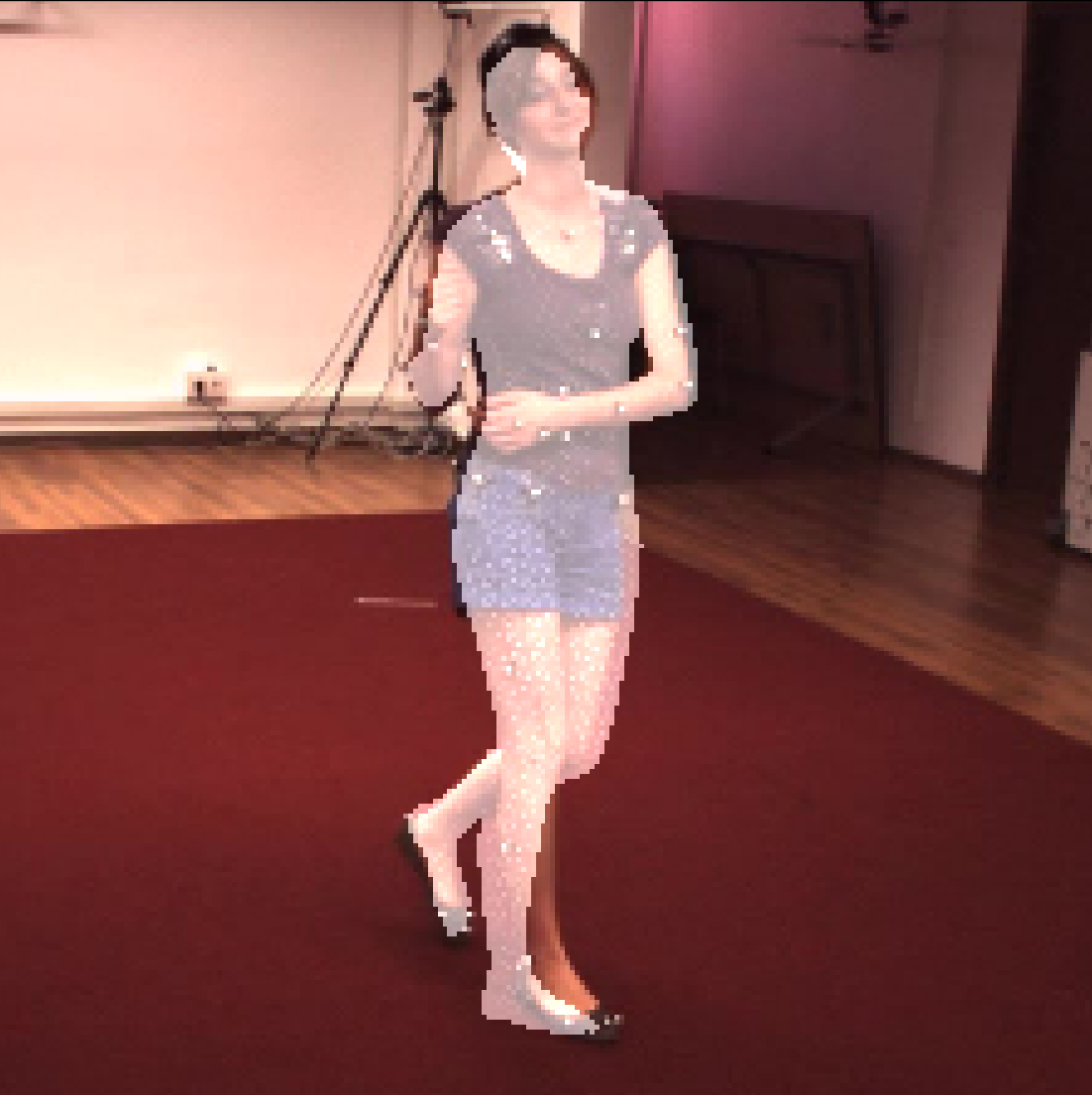}
    \includegraphics[width=\qualfigwidth\textwidth]{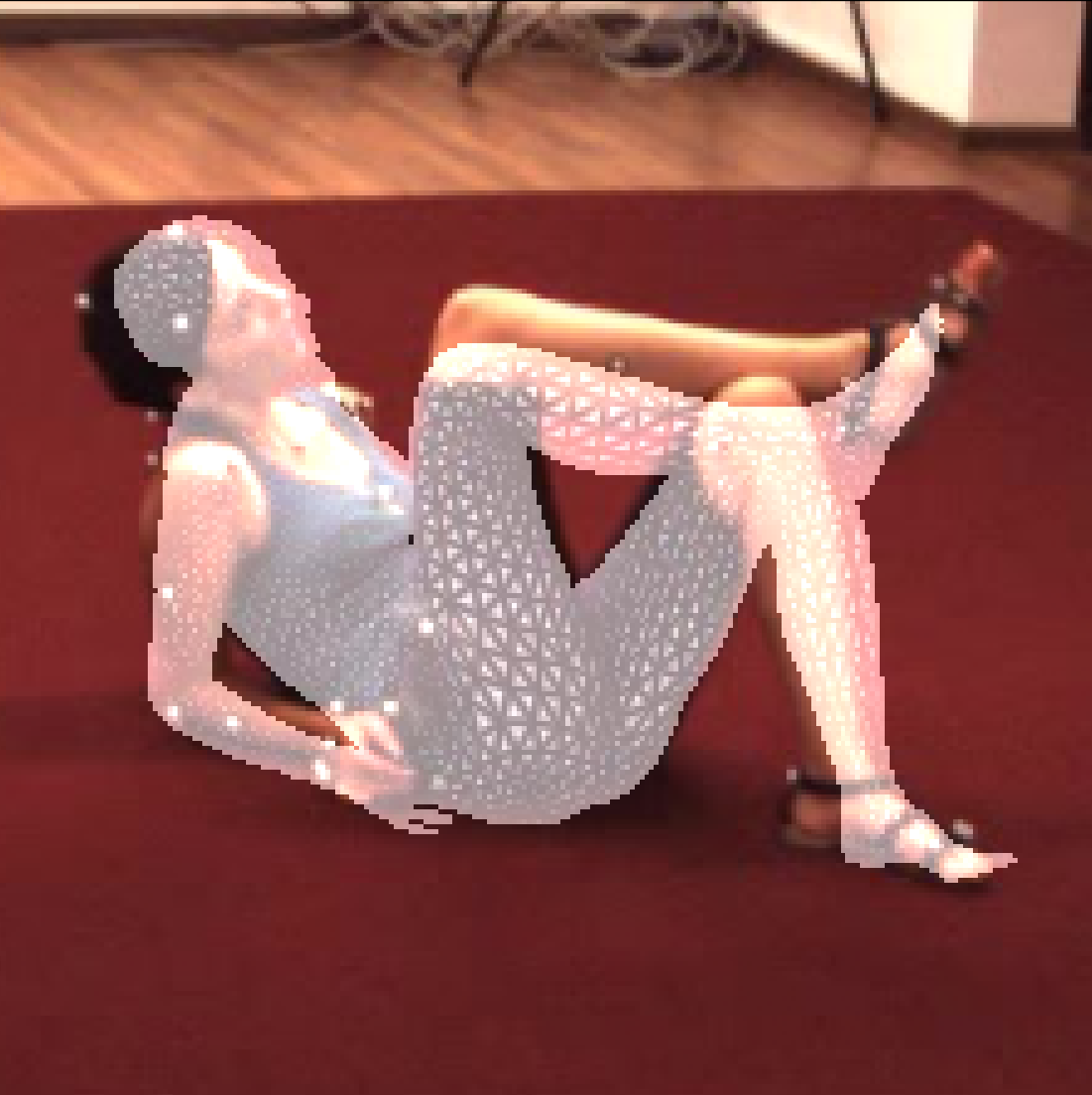}
    \includegraphics[width=\qualfigwidth\textwidth]{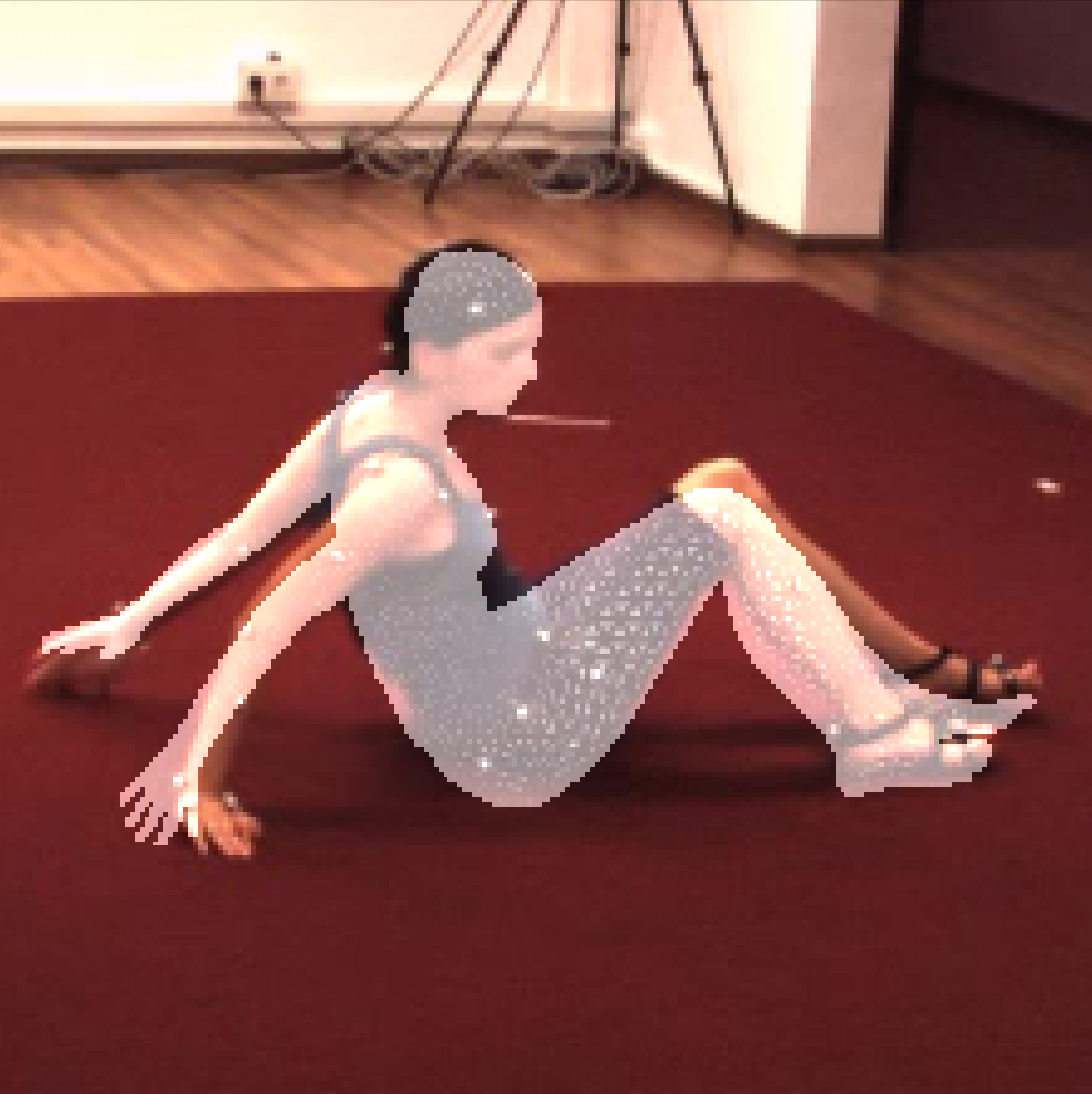}
    \includegraphics[width=\qualfigwidth\textwidth]{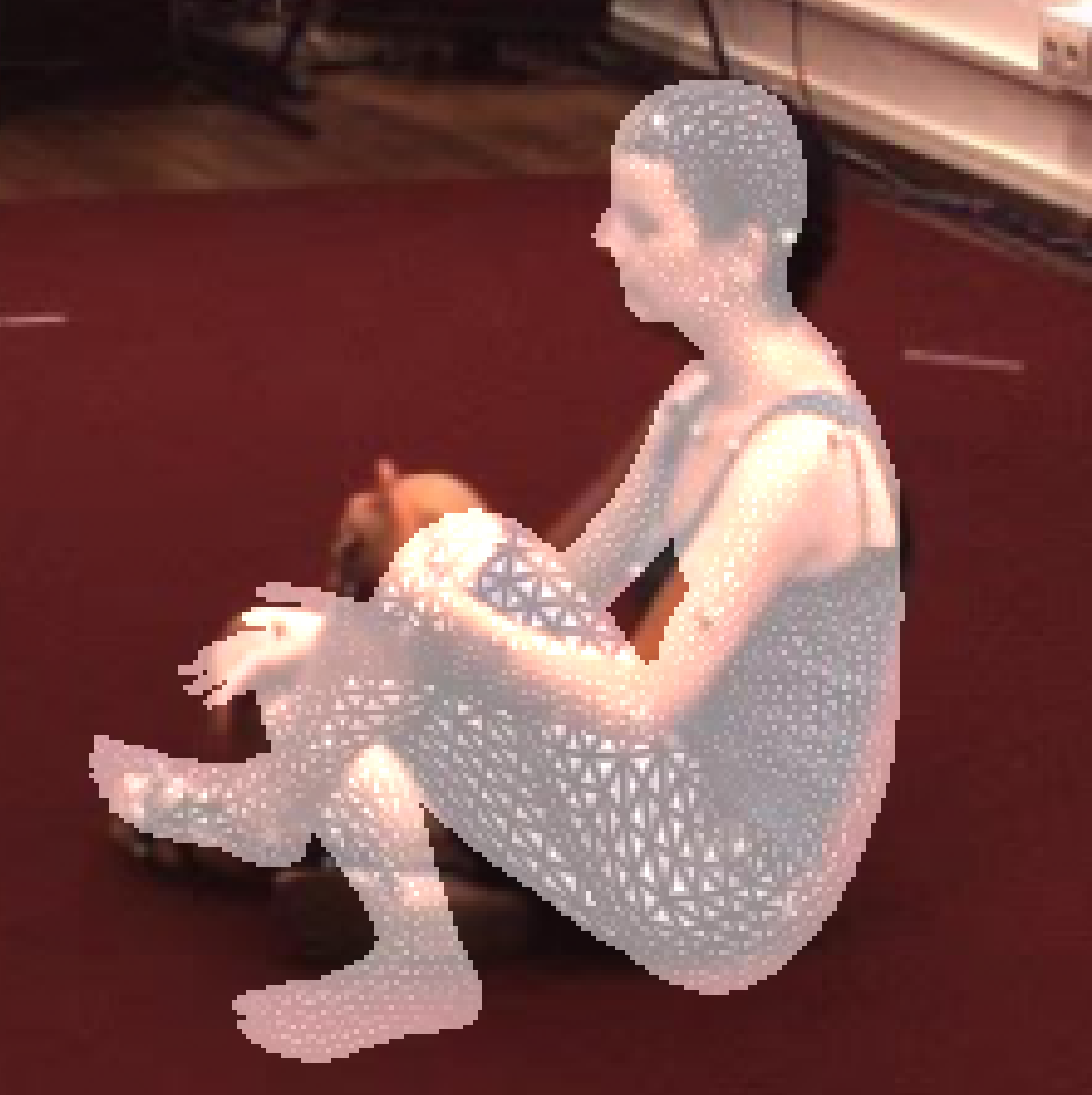}
    \includegraphics[width=\qualfigwidth\textwidth]{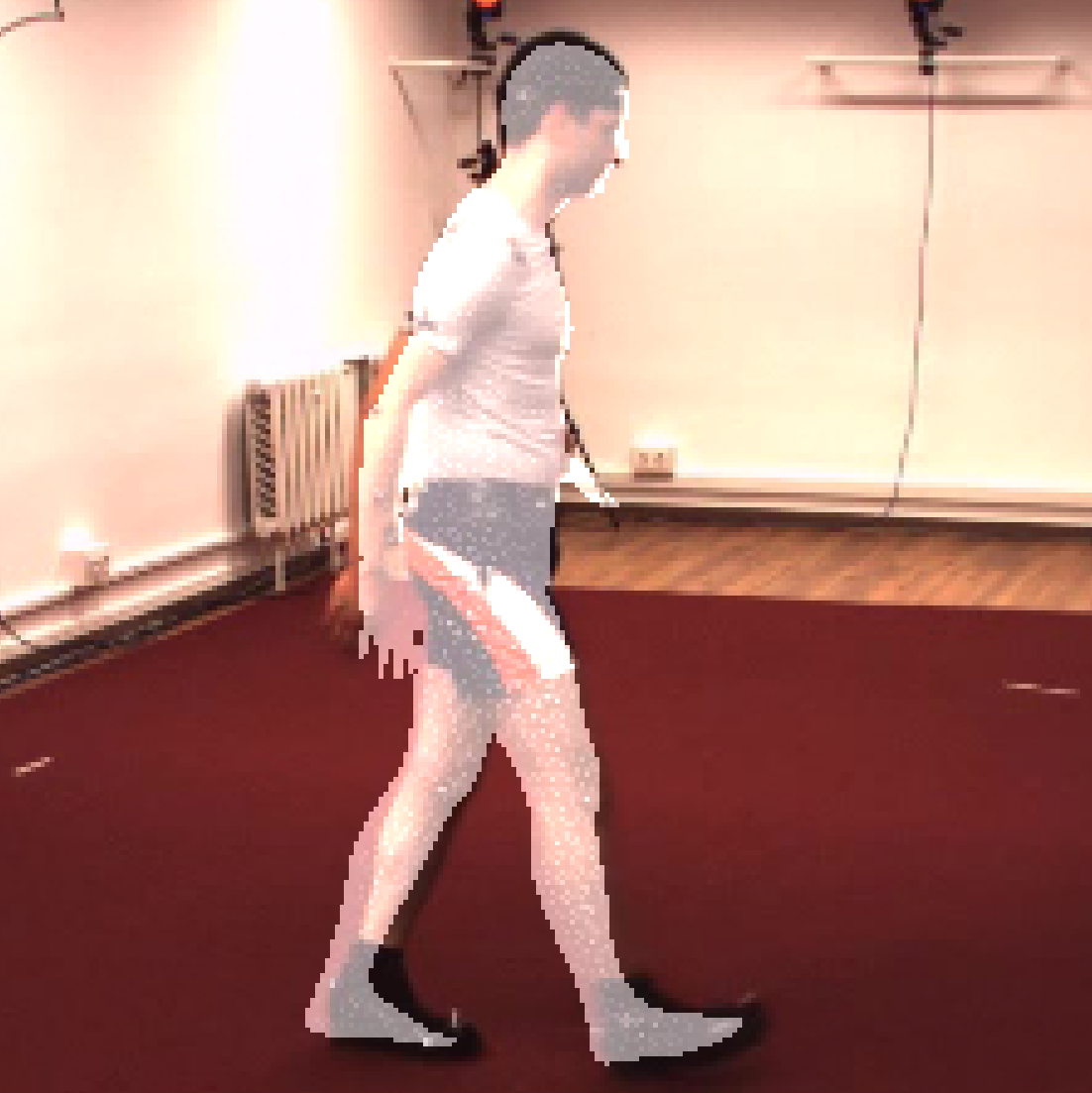}
    \\
    \vspace{0.3em}
    \includegraphics[width=\qualfigwidth\textwidth]{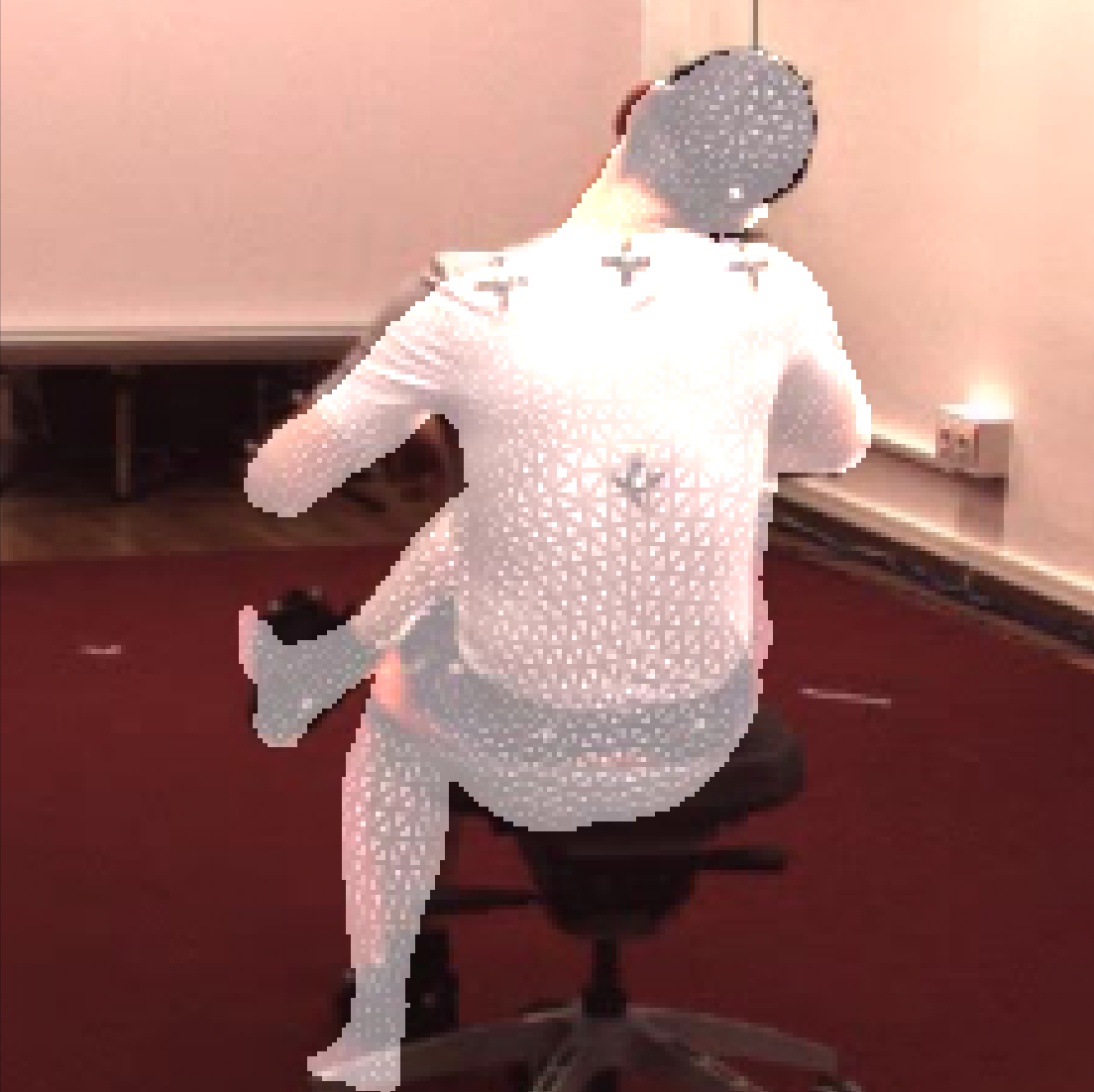}
    \includegraphics[width=\qualfigwidth\textwidth]{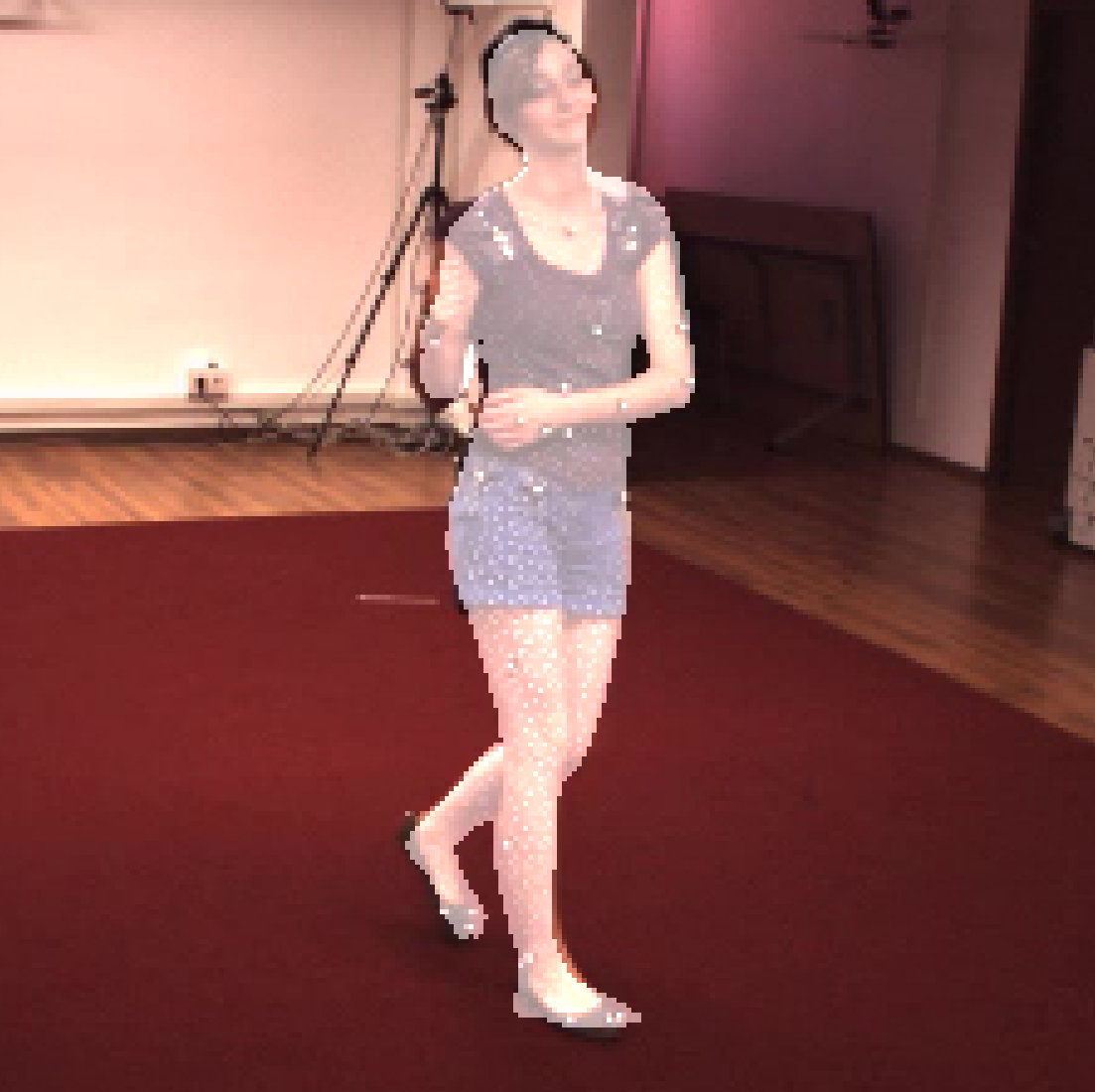}
    \includegraphics[width=\qualfigwidth\textwidth]{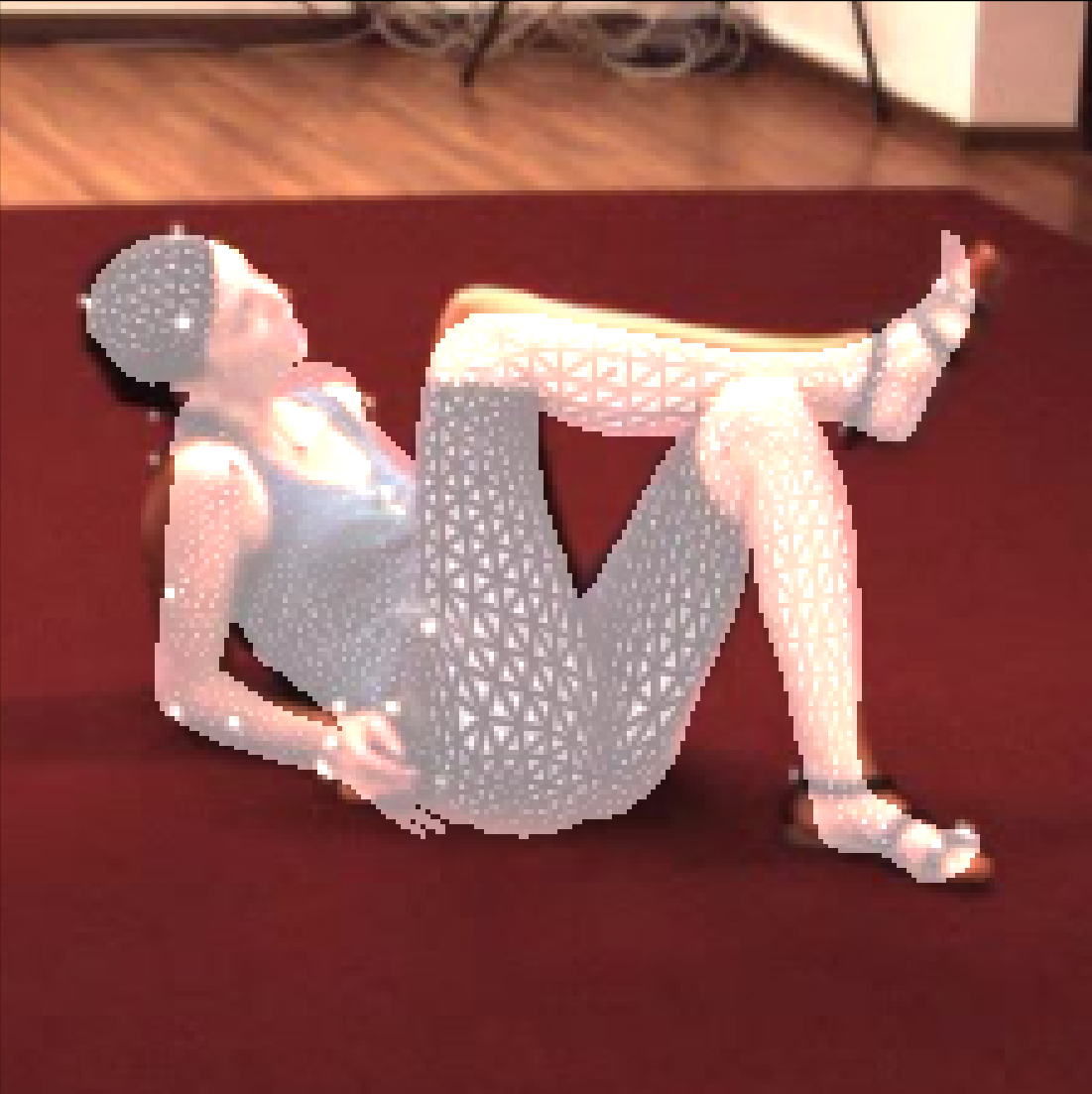}
    \includegraphics[width=\qualfigwidth\textwidth]{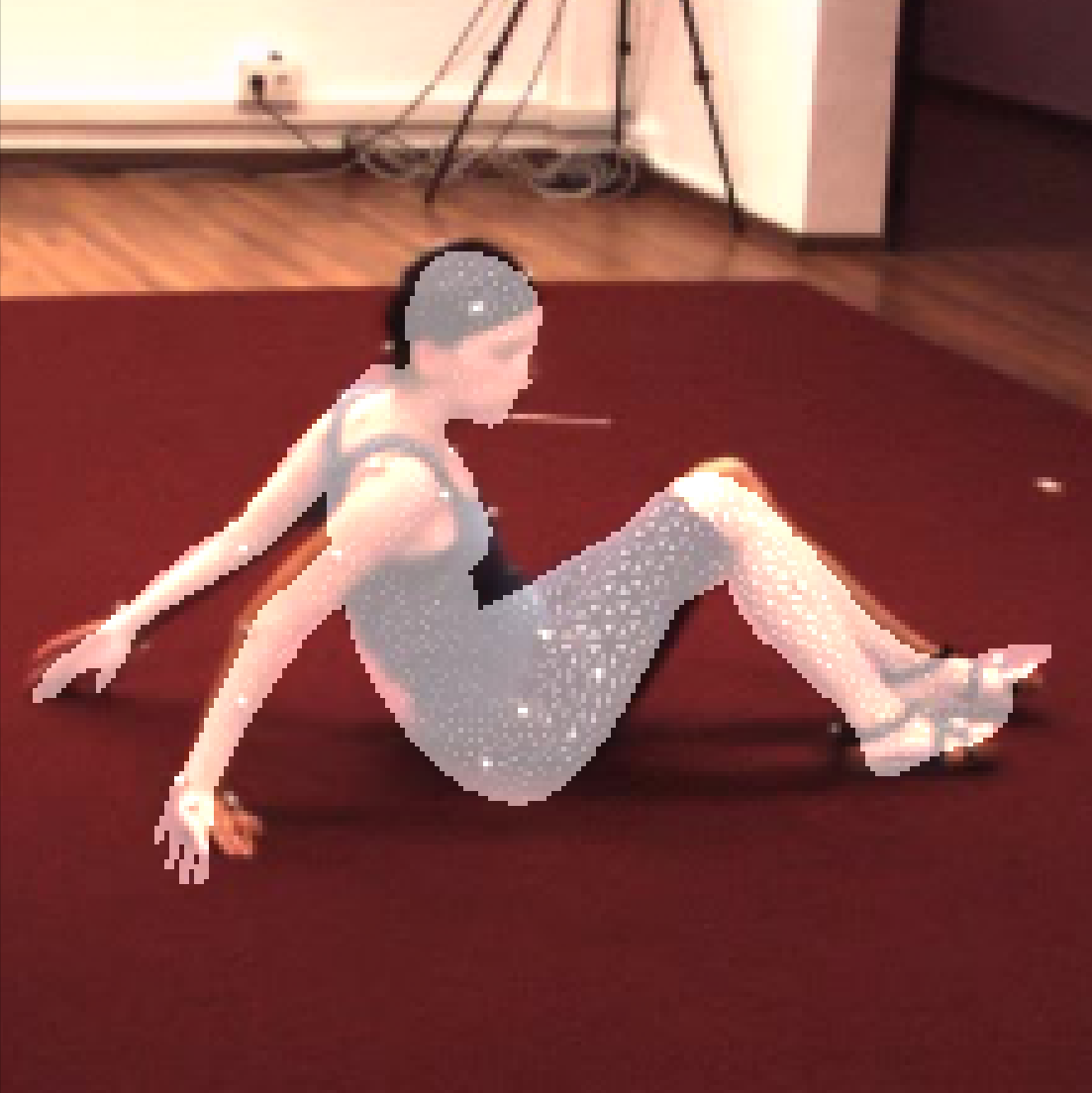}
    \includegraphics[width=\qualfigwidth\textwidth]{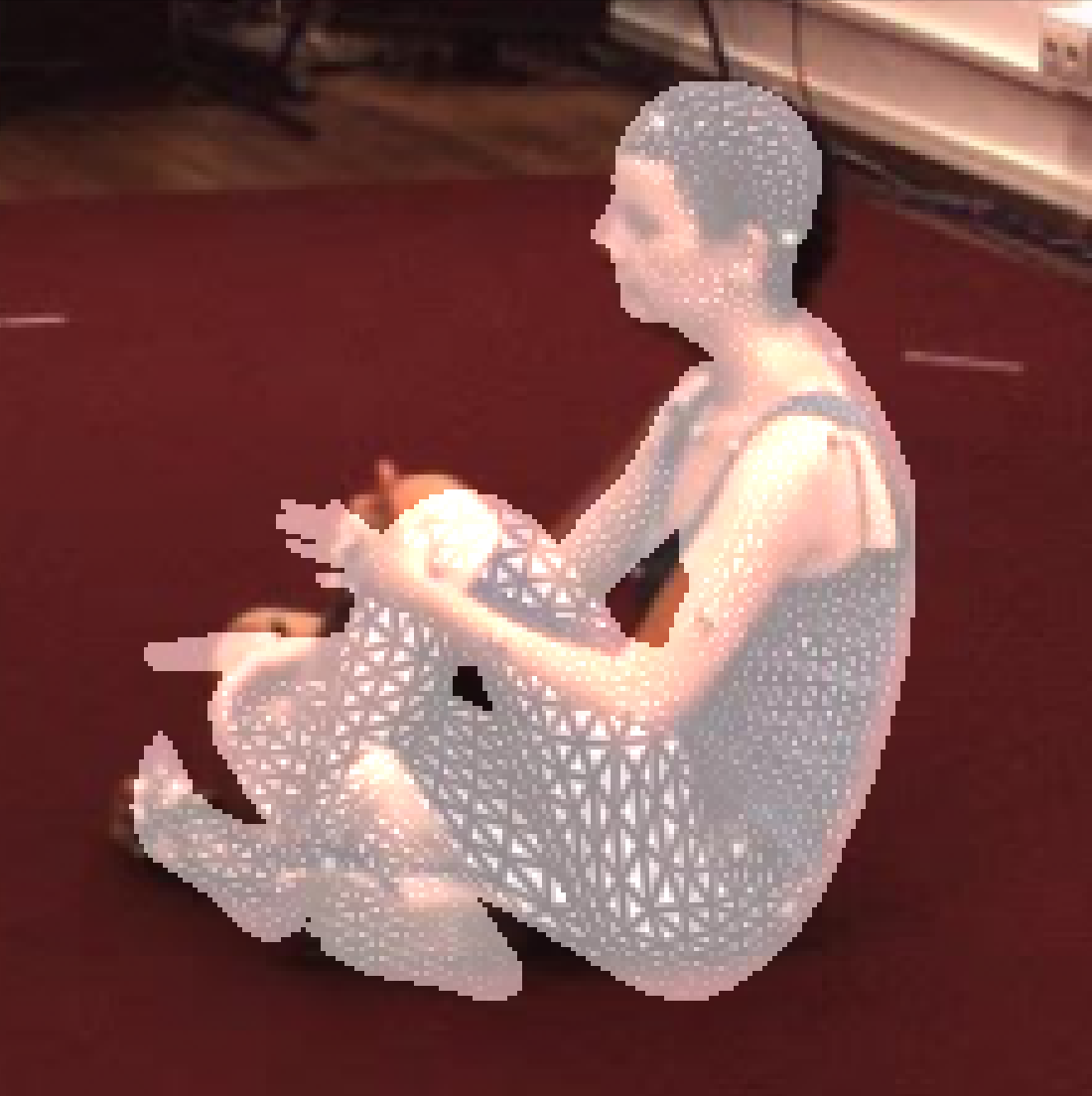}
    \includegraphics[width=\qualfigwidth\textwidth]{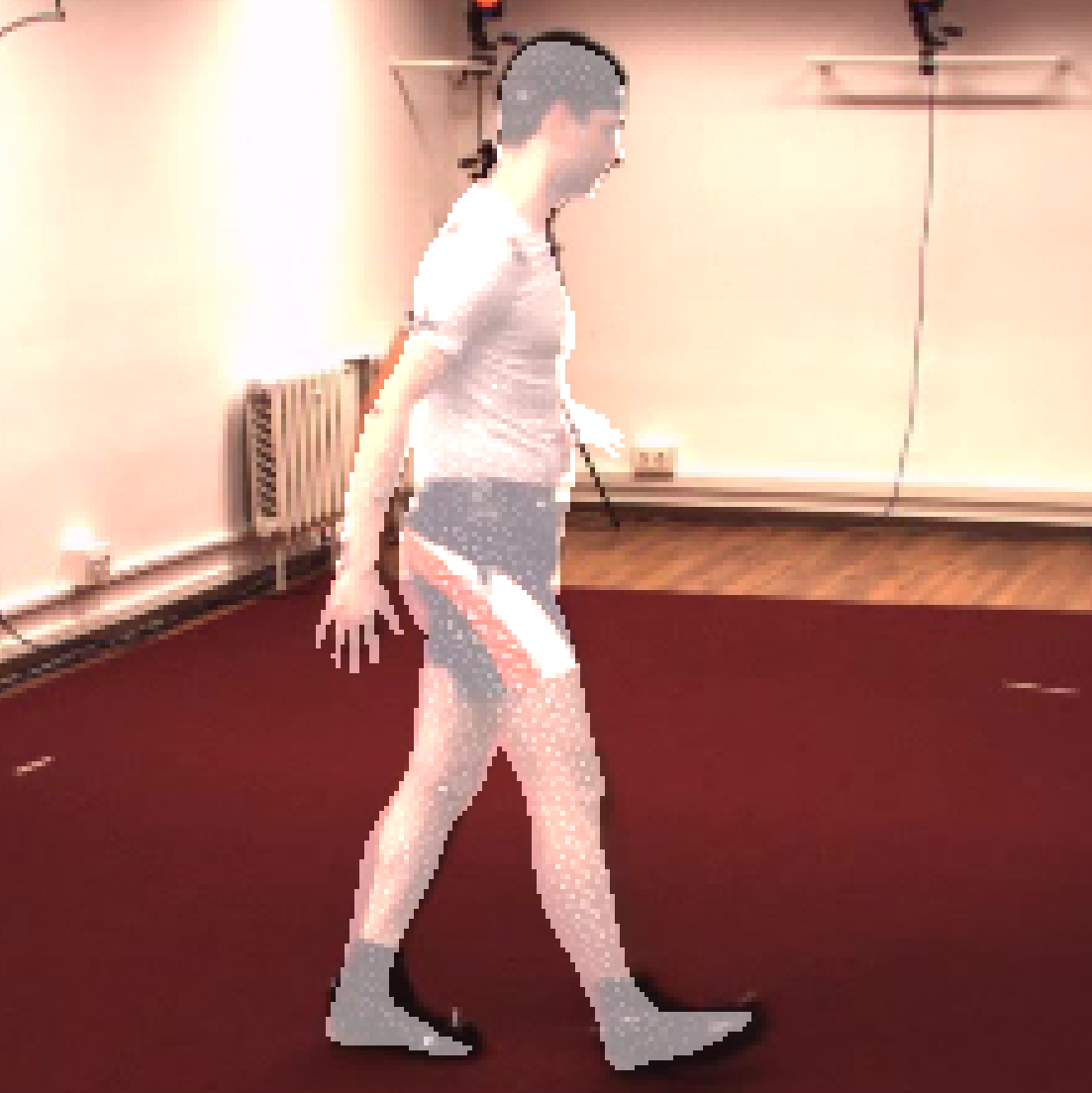}
    \caption{
    {\bf Qualitative examples -- }
    (Top row) initial estimates by SPIN and (bottom row) our meshized pseudo ground-truth SMPL parameters.
    }
    \label{fig:qualitative_mesh}%
\end{figure*}

\subsection{The pseudo ground truth}

We show qualitative examples of our pseudo ground truth and the standard SMPL parameters we start with in \Figure{qualitative_mesh}. 
As shown, the poses are realistic and fit the subject's silhouette much better than the standard SPIN estimates. 

We show in \Figure{ablation_discrim} that without a discriminator, the poses can break. 
This fragility is because it is often the case that the quickest way for a pose to be satisfied is through unrealistic contortions of the body. 

\begin{figure}%
    \centering
    \includegraphics[width=0.49\linewidth]{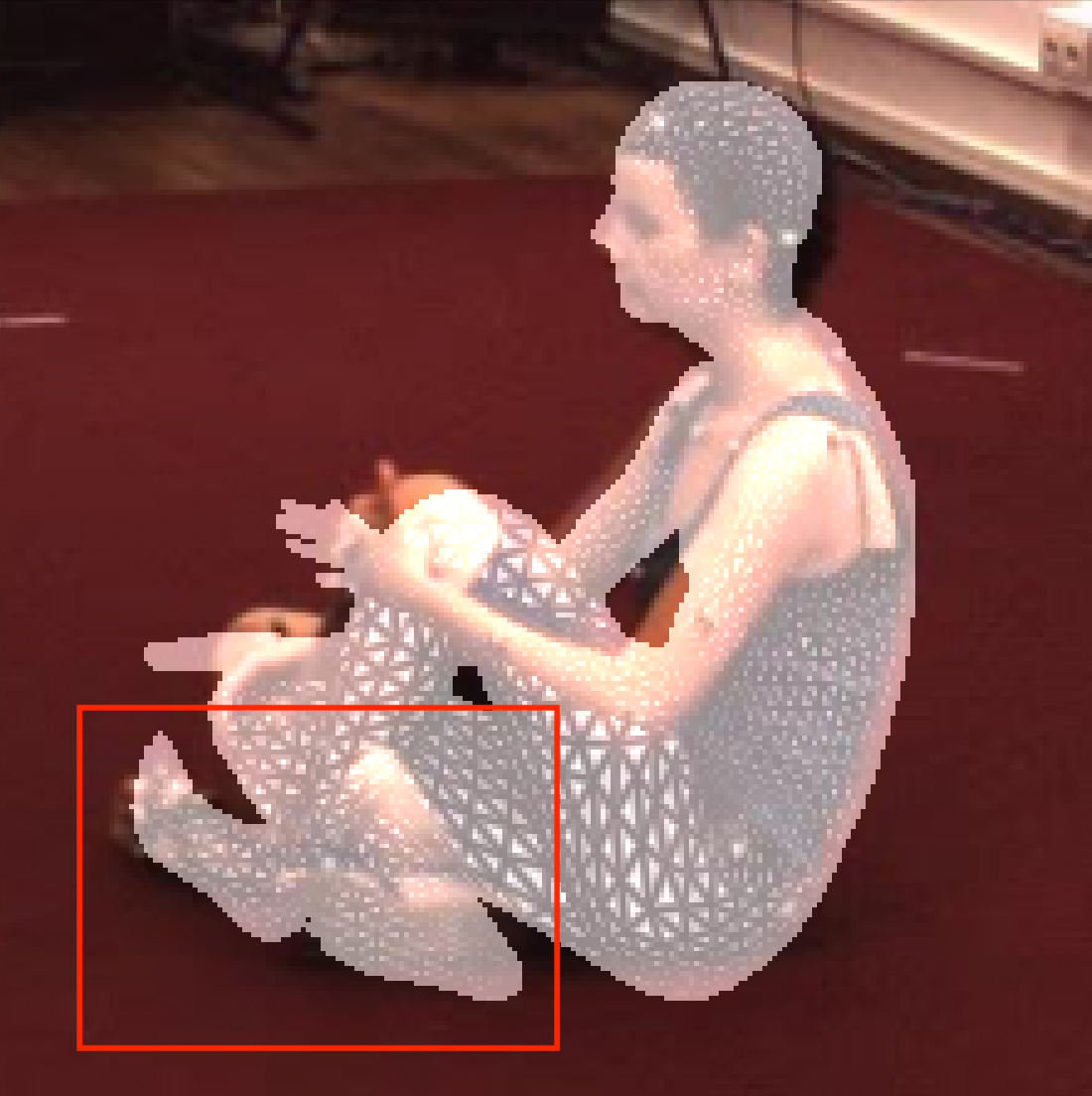}
    \includegraphics[width=0.49\linewidth]{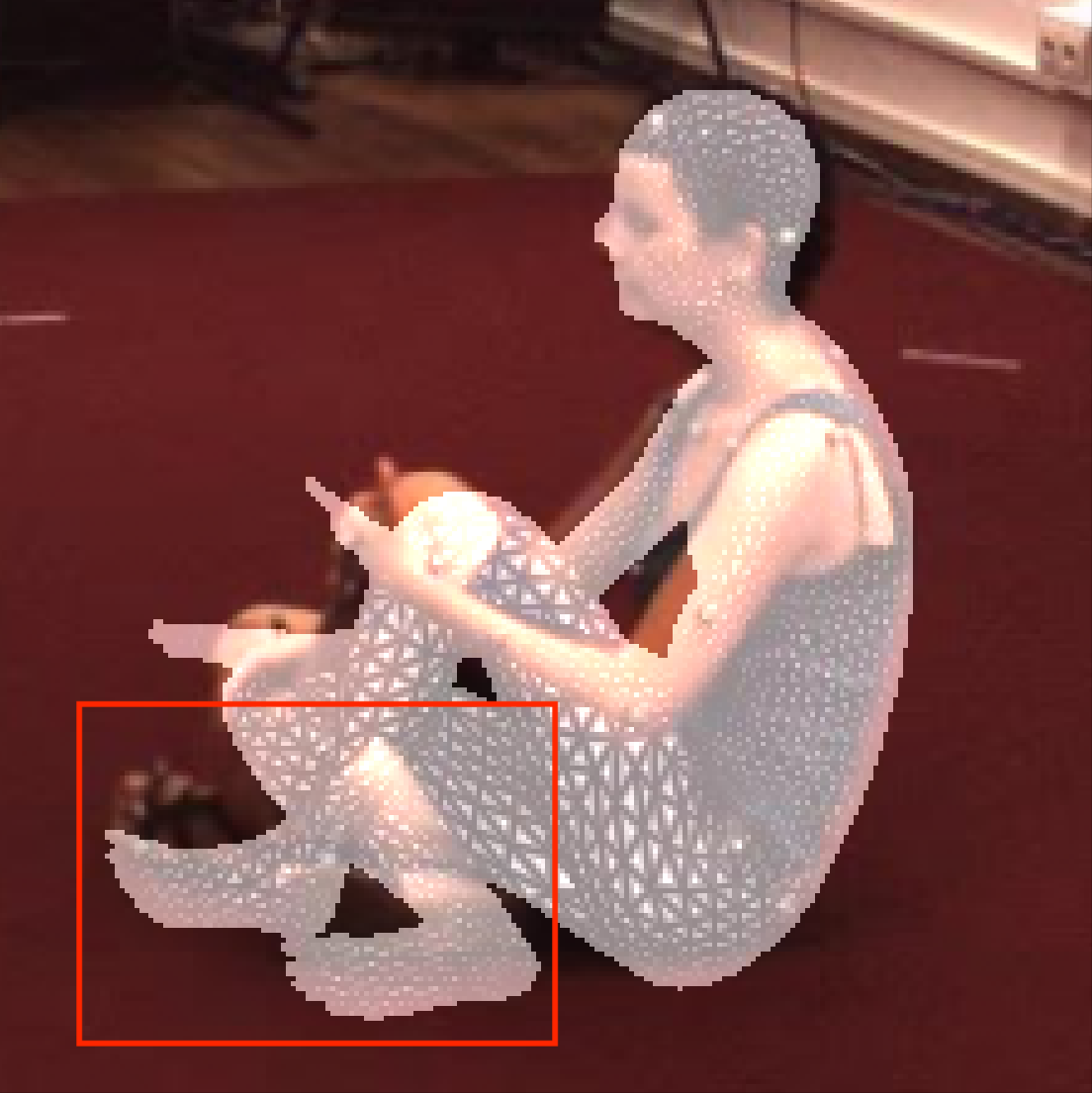}
    \caption{
    {\bf Silhouette ablation -- }
    We show the optimized SMPL parameters (left) with and (right) without the silhouette loss, overlaid onto the image.
    Having the silhouette loss encourages the estimated pose match the silhouette of the subject in the scene.
    }%
    \label{fig:opt_without_silhouette}%
\end{figure}

Nothing encourages the pose to align with the human subject without the silhouette term. 
\Figure{opt_without_silhouette} shows one example of this is necessary. 
Only the heel joint supervises the location of the foot. 
This lack of supervision means that the foot can point in any direction while still satisfying this joint. 
Only the silhouette term encourages the toes to lie within the body. 

Improvements brought on by minimizing the energy related to the silhouette show that it is important to consider the image when estimating the pseudo-ground-truth poses.

\subsection{The joint regressor}

In \Table{jreg_results} we report the performances of SPIN~\cite{spin}, VIBE~\cite{vibe}, and MEVA~\cite{meva} with our improved joint regressors and the standard regressor.
First, we show that simply re-training the regressor from SPIN predictions to H3.6M ground truth significantly improves results, indicating that a mismatching joint regressor was used for training SPIN and related methods~\cite{spin, vibe, meva} across different datasets. 
Our full joint regressor trained on the refined poses further improves results. 
This improvement is only subtle as the refined $J_\text{spin}$ already outputs 3D joint locations consistent with the H3.6M ground truth but with a misaligned body shape.


Applying the improved joint regressor to other methods (VIBE and MEVA) also improves their results. 
Since this can be seen across a set of methods, it indicates that the joint regressor $\regressor_\text{opt}$ is not learning something specific to SPIN but to all models that utilize the SMPL mesh.

\begin{figure}%
    \centering
    \includegraphics[width=0.49\linewidth]{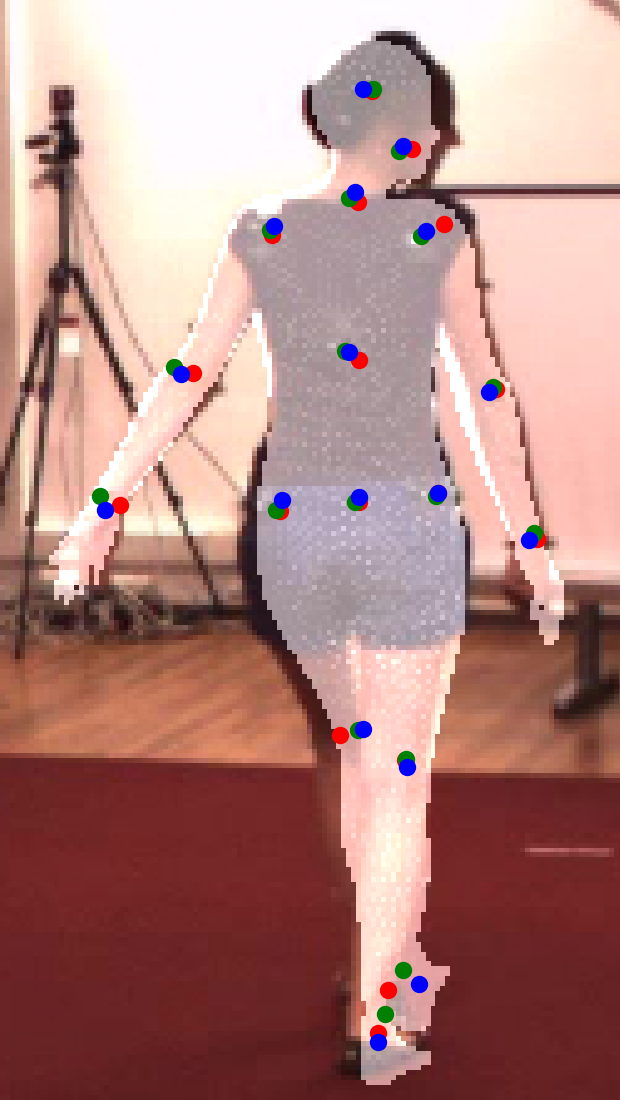}
    \includegraphics[width=0.49\linewidth]{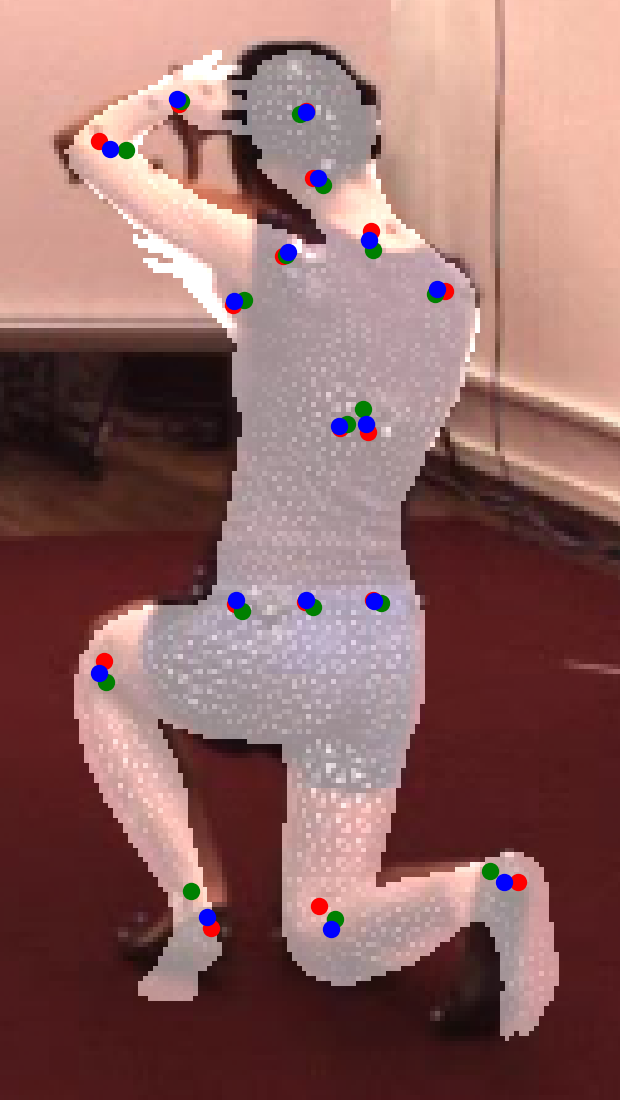}
    \caption{
    {\bf Joint estimation examples -- }
    Example joint estimates from the same estimated SMPL pose with with \textit{different} SMPL-to-joint regressors.
    Ground-truth joint locations are marked with \textcolor{red}{red}, projected joints from the commonly used joint regressor $\regressor_\text{std}$ in \textcolor{dark_green}{green}, and projected joints from our joint regressor $\regressor_\text{opt}$ in  \textcolor{blue}{blue}.
    }%
    \label{fig:joint_est}%
\end{figure}

As can be seen in \Figure{joint_est}, the new joint regressor matches the ground truth joints better compared to the standard joint regressor.



\begin{table}[]
\resizebox{\linewidth}{!}{
\setlength{\tabcolsep}{2pt}
\begin{tabular}{lcc cc cc}
    \toprule
     & \multicolumn{2}{c}{\begin{tabular}[c]{@{}l@{}}Current joint \\ regressor $\regressor_\text{std}$\end{tabular}} & \multicolumn{2}{c}{\begin{tabular}[c]{@{}l@{}}Retrained joint \\ regressor $\regressor_\text{spin}$\end{tabular}} & \multicolumn{2}{c}{\begin{tabular}[c]{@{}l@{}}Optimized joint \\ regressor $\regressor_\text{opt}$\end{tabular}} \\
     \cmidrule(lr){2-3} 
     \cmidrule(lr){4-5} 
     \cmidrule(lr){6-7} 
     & MPJPE $\downarrow$                & \begin{tabular}[c]{@{}l@{}}PA-MPJPE $\downarrow$\end{tabular}               & MPJPE $\downarrow$                & \begin{tabular}[c]{@{}l@{}}PA-MPJPE $\downarrow$\end{tabular}                & MPJPE $\downarrow$                & \begin{tabular}[c]{@{}l@{}}PA-MPJPE $\downarrow$\end{tabular}               \\ 
     \midrule
SPIN & 60.4                 & 41.3                                                              & 57.0                 & \textbf{39.6}                                                               & \textbf{56.9}                 & 39.7                                                              \\
VIBE & 62.9                 & 43.1                                                              & 60.1                 & 41.7                                                               & \textbf{60.0}                 & \textbf{41.5}                                                              \\
MEVA & 101.6                & 62.3                                                              & 98.4                 & \textbf{61.9}                                                               & \textbf{98.3}                 & \textbf{61.9}                                                      
\\ 
\bottomrule
\end{tabular}
}
\caption{Without retraining, these methods improve MPJPE just by using the new joint regressor $\regressor_\text{opt}$. 
Each of these methods was run in single-frame mode, which would hurt the performance of VIBE and MEVA. 
Highlighted in bold are the results per model that performed the best. MPJPE and PA-MPJPE are respectively unaligned and procrustes aligned mean per joint position errors shown in millimeters.  }
\label{tab:jreg_results}
\end{table}


We believe that retraining these methods \cite{spin, vibe, meva} from scratch with the optimized joint regressor $\regressor_\text{opt}$ would lead to better shape alignment. 
This is because the optimized joint regressor best matches the points on the SMPL mesh with their corresponding ground truth joints and image evidence.
Training on Human3.6m with the current joint regressor would lead to a consistent shift in the poses from ground truth. 
As a result, we suggest that future human pose estimation methods training on Human3.6m use our pseudo-ground-truth poses and retrained joint regressor.
\footnote{The code will be made public at https://github.com/ubc-vision/human-body-pose.}

\subsection{Limitations and discussions}

\begin{figure}%
    \centering
    \subfloat[\centering An example of an incorrect ankle label]{{\includegraphics[width=0.4\textwidth]{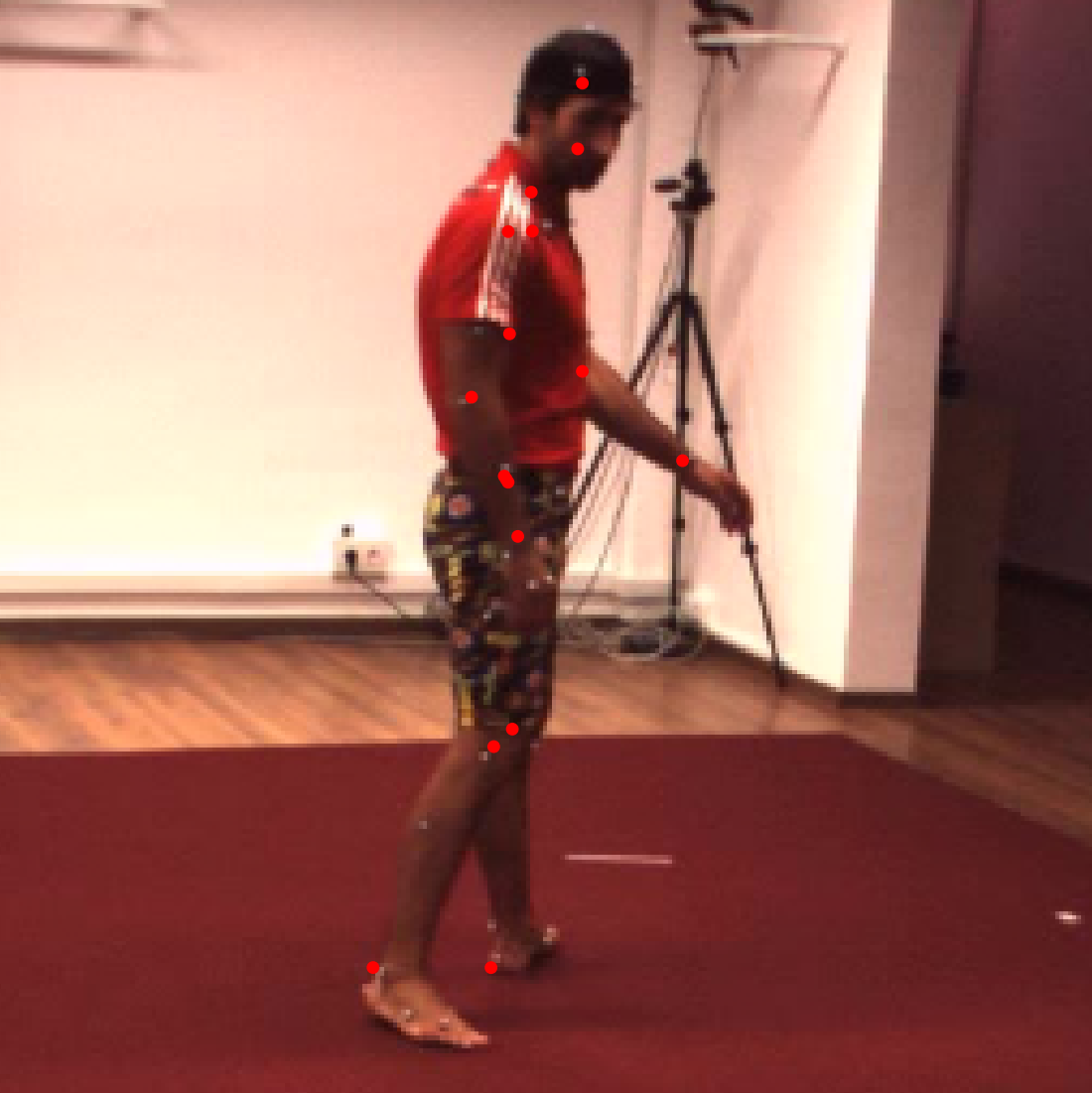} }}\\%
    \subfloat[\centering An example of accurate ground truth labels]{{\includegraphics[width=0.4\textwidth]{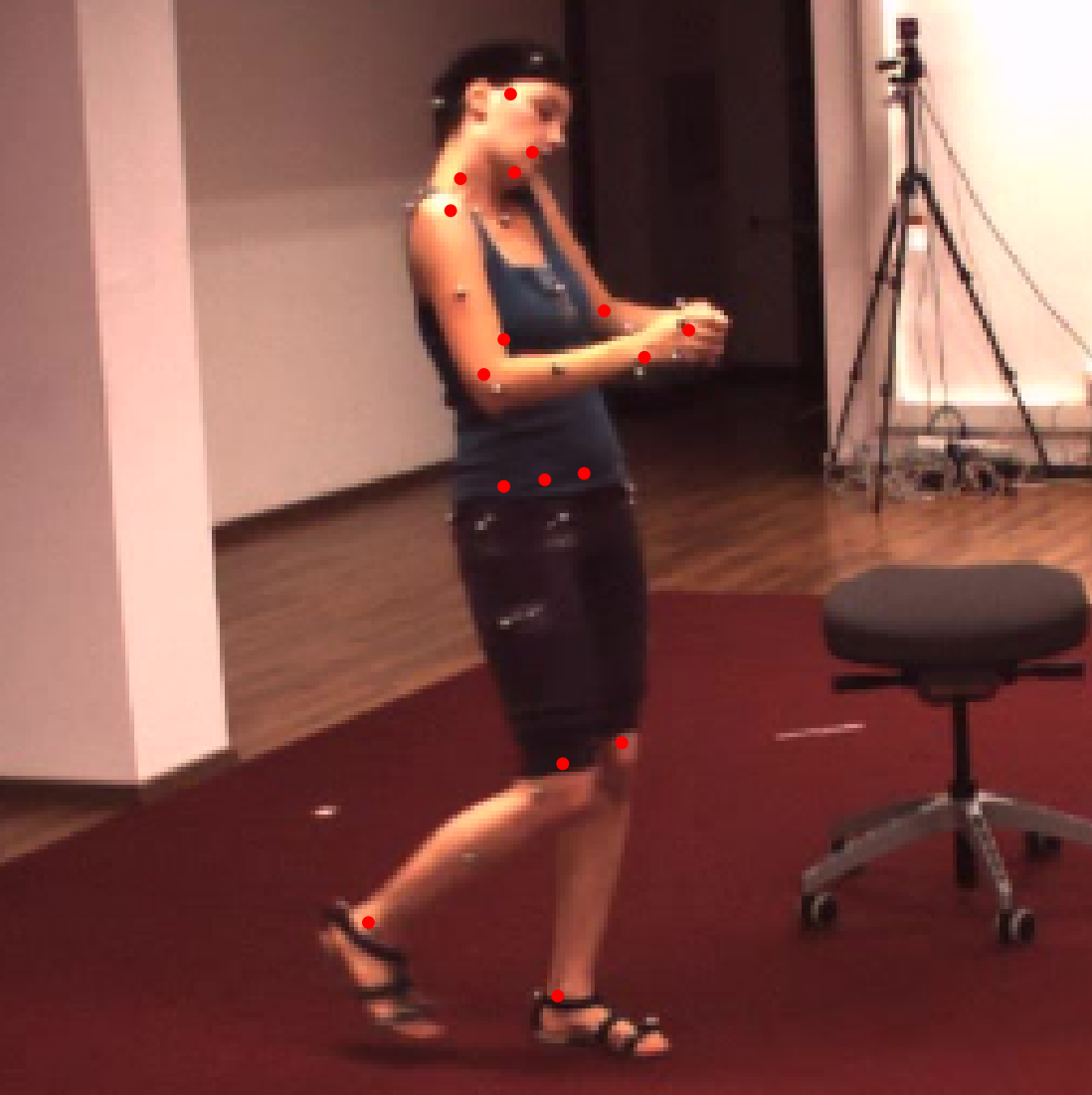} }}\\%
    \caption{ Here is a visualization of the initial image and their corresponding overlaid ground truth 2D joints visualized in \textcolor{red}{red}. When looking at the ankles, it is evident that these subjects are labelled differently. The subject on the left has his ground truth right ankle behind his right foot, and his ground truth left ankle is below his left foot. The subject on the bottom seems to have been correctly labelled. }%
    \label{fig:gt_j2d_comp}%
\end{figure}

While our method delivers pseudo ground truth that can be used to achieve better joint regression, there still exists error between the regressed joints using our pseudo ground truth $\regressor_\text{opt}\smpl(\poseNshape_\text{opt})$ and the annotated joint locations $\gtpoints$.
This error is partly because of the label inconsistency across different subjects; See~\Figure{gt_j2d_comp}.
Inconsistent labelling between subjects goes against the basic assumption of the joint regressor---only one joint regressor should exist for all subjects.
We note that the 2D and 3D joint locations are defined purely from ground-truth annotations (camera and joint labels) and cannot be improved unless data is re-captured.
Correcting for this error is beyond the scope of our research but is a worthwhile direction of investigation.

\section{Conclusion}

We have shown that one can improve the state-of-the-art in human pose estimation by simply  improving the joint regressor.
We empirically demonstrated that the current widely accepted joint regressor can create error even when the SMPL estimates are accurate.
We thus proposed a pseudo ground truth generation method based on aligning SMPL meshes with the target subject's silhouette, as well as enforcing plausible poses via an adversarial setup.
We then demonstrated that this leads to an improved joint regressor, which then leads to improved pose estimation.
This indicates that the performance of the SMPL-based pose estimation methods is likely under-reported in the literature.

%
\IEEEpeerreviewmaketitle

\section*{Acknowledgment}
The authors would like to thank Wei Jiang for his helpful insights. 
This work was partly supported by compute Canada.



\bibliographystyle{IEEEtran}
\bibliography{IEEEabrv,macros.bib,biblio}

\begin{thebibliography}{10}
\providecommand{\url}[1]{#1}
\csname url@samestyle\endcsname
\providecommand{\newblock}{\relax}
\providecommand{\bibinfo}[2]{#2}
\providecommand{\BIBentrySTDinterwordspacing}{\spaceskip=0pt\relax}
\providecommand{\BIBentryALTinterwordstretchfactor}{4}
\providecommand{\BIBentryALTinterwordspacing}{\spaceskip=\fontdimen2\font plus
\BIBentryALTinterwordstretchfactor\fontdimen3\font minus
  \fontdimen4\font\relax}
\providecommand{\BIBforeignlanguage}[2]{{%
\expandafter\ifx\csname l@#1\endcsname\relax
\typeout{** WARNING: IEEEtran.bst: No hyphenation pattern has been}%
\typeout{** loaded for the language `#1'. Using the pattern for}%
\typeout{** the default language instead.}%
\else
\language=\csname l@#1\endcsname
\fi
#2}}
\providecommand{\BIBdecl}{\relax}
\BIBdecl

\bibitem{dantone2013human}
M.~Dantone, J.~Gall, C.~Leistner, and L.~Van~Gool, ``{Human pose estimation
  using body parts dependent joint regressors},'' in \emph{Conference on
  Computer Vision and Pattern Recognition}, 2013.

\bibitem{gkioxari2014using}
G.~Gkioxari, B.~Hariharan, R.~Girshick, and J.~Malik, ``{Using k-poselets for
  detecting people and localizing their keypoints},'' in \emph{Conference on
  Computer Vision and Pattern Recognition}, 2014.

\bibitem{sminchisescu2002human}
C.~Sminchisescu and A.~C. Telea, ``Human pose estimation from silhouettes. a
  consistent approach using distance level sets,'' in \emph{10th International
  Conference on Computer Graphics, Visualization and Computer Vision
  (WSCG'02)}, vol.~10, 2002.

\bibitem{hmr}
A.~Kanazawa, M.~J. Black, D.~W. Jacobs, and J.~Malik, ``{End-to-end recovery of
  human shape and pose},'' in \emph{Conference on Computer Vision and Pattern
  Recognition}, 2018.

\bibitem{smpl}
M.~Loper, N.~Mahmood, J.~Romero, G.~Pons-Moll, and M.~J. Black, ``{SMPL: A
  skinned multi-person linear model},'' \emph{ACM Transactions on Graphics},
  vol.~34, no.~6, pp. 1--16, 2015.

\bibitem{vibe}
M.~Kocabas, N.~Athanasiou, and M.~J. Black, ``{Vibe: Video inference for human
  body pose and shape estimation},'' in \emph{Conference on Computer Vision and
  Pattern Recognition Workshops}, 2020.

\bibitem{spin}
N.~Kolotouros, G.~Pavlakos, M.~J. Black, and K.~Daniilidis, ``{Learning to
  reconstruct 3D human pose and shape via model-fitting in the loop},'' in
  \emph{International Conference on Computer Vision Workshops}, 2019.

\bibitem{human3.6m}
C.~Ionescu, D.~Papava, V.~Olaru, and C.~Sminchisescu, ``{Human3. 6m: Large
  scale datasets and predictive methods for 3d human sensing in natural
  environments},'' \emph{IEEE Transactions on Pattern Analysis and Machine
  Intelligence}, vol.~36, no.~7, pp. 1325--1339, 2013.

\bibitem{survey}
C.~Zheng, W.~Wu, T.~Yang, S.~Zhu, C.~Chen, R.~Liu, J.~Shen, N.~Kehtarnavaz, and
  M.~Shah, ``{Deep learning-based human pose estimation: A survey},''
  \emph{arXiv Preprint}, 2020.

\bibitem{3dpw}
T.~von Marcard, R.~Henschel, M.~Black, B.~Rosenhahn, and G.~Pons-Moll,
  ``{Recovering Accurate 3D Human Pose in The Wild Using IMUs and a Moving
  Camera},'' in \emph{European Conference on Computer Vision}, 2018.

\bibitem{total_capture}
H.~Joo, T.~Simon, and Y.~Sheikh, ``{Total capture: A 3d deformation model for
  tracking faces, hands, and bodies},'' in \emph{Conference on Computer Vision
  and Pattern Recognition Workshops}, 2018.

\bibitem{2ddataset}
M.~Andriluka, L.~Pishchulin, P.~Gehler, and B.~Schiele, ``{2D Human Pose
  Estimation: New Benchmark and State of the Art Analysis},'' in
  \emph{Conference on Computer Vision and Pattern Recognition}, 2014.

\bibitem{IonescuSminchisescu11}
C.~S. Catalin~Ionescu, Fuxin~Li, ``Latent structured models for human pose
  estimation,'' in \emph{International Conference on Computer Vision}, 2011.

\bibitem{mosh}
M.~Loper, N.~Mahmood, and M.~J. Black, ``{MoSh: Motion and shape capture from
  sparse markers},'' \emph{ACM Transactions on Graphics}, vol.~33, no.~6, pp.
  1--13, 2014.

\bibitem{scape}
D.~Anguelov, P.~Srinivasan, D.~Koller, S.~Thrun, J.~Rodgers, and J.~Davis,
  ``Scape: shape completion and animation of people,'' in \emph{ACM SIGGRAPH},
  2005.

\bibitem{maskrcnn}
K.~He, G.~Gkioxari, P.~Doll{\'a}r, and R.~Girshick, ``{Mask r-cnn},'' in
  \emph{International Conference on Computer Vision}, 2017.

\bibitem{pytorch3d}
N.~Ravi, J.~Reizenstein, D.~Novotny, T.~Gordon, W.-Y. Lo, J.~Johnson, and
  G.~Gkioxari, ``Accelerating 3d deep learning with pytorch3d,''
  \emph{arXiv:2007.08501}, 2020.

\bibitem{metro}
K.~Lin, L.~Wang, and Z.~Liu, ``{End-to-end human pose and mesh reconstruction
  with transformers},'' in \emph{Conference on Computer Vision and Pattern
  Recognition}, 2021.

\bibitem{lifting}
D.~Tome, C.~Russell, and L.~Agapito, ``{Lifting from the deep: Convolutional 3d
  pose estimation from a single image},'' in \emph{Conference on Computer
  Vision and Pattern Recognition}, 2017.

\bibitem{pavlakos2019expressive}
G.~Pavlakos, V.~Choutas, N.~Ghorbani, T.~Bolkart, A.~A. Osman, D.~Tzionas, and
  M.~J. Black, ``{Expressive body capture: 3d hands, face, and body from a
  single image},'' in \emph{Conference on Computer Vision and Pattern
  Recognition Workshops}, 2019.

\bibitem{smplx}
------, ``{Expressive body capture: 3d hands, face, and body from a single
  image},'' in \emph{Conference on Computer Vision and Pattern Recognition},
  2019.

\bibitem{meva}
Z.~Luo, S.~A. Golestaneh, and K.~M. Kitani, ``3d human motion estimation via
  motion compression and refinement,'' 2020.

\bibitem{lawson1995solving}
C.~L. Lawson and R.~J. Hanson, \emph{{Solving least squares problems}}.\hskip
  1em plus 0.5em minus 0.4em\relax SIAM, 1995.

\bibitem{gan}
I.~Goodfellow, J.~Pouget-Abadie, M.~Mirza, B.~Xu, D.~Warde-Farley, S.~Ozair,
  A.~Courville, and Y.~Bengio, ``{Generative adversarial nets},'' in
  \emph{Advances in Neural Information Processing Systems}, 2014.

\bibitem{adam}
D.~P. Kingma and J.~Ba, ``Adam: A method for stochastic optimization,''
  \emph{International Conference on Learning Representations}, 2014.

\end{thebibliography}
%



\end{document}